
\def\ignore#1{}
 

\newcount\sectnum
\newcount\subsectnum
\newcount\eqnumber

\global\eqnumber=1\sectnum=0


\def\lab{(\the\sectnum.\the\eqnumber)}



\def\show#1{#1}



\def\smskip{\vskip 5 pt}
\def\medskip{\vskip 10 pt}
\def\bigskip{\vskip 15 pt}
\def\pn{\par\noindent}

\def\frac#1#2{{#1\over #2}}

\def\a{\alpha}

\def\m{\mu}
\def\p{\pi}

\def\P{\Pi}

\def\tl{\tilde}

\def\old#1{}
\def\leaderfill{\leaders\hbox to 1em{\hss.\hss}\hfill}


\parindent=2pc
\baselineskip=15pt
\vsize=8.7 true in
\voffset=0.125 true in
\parskip=3pt


\def\minprob#1#2#3{$$\eqalign{&\hbox{minimize\ \ }#1\cr &\hbox{subject to\ \
}#2\cr}\ifnum 0=#3{}\else\eqno(#3)\fi$$}        
     
\def\maxprob#1#2#3{$$\eqalign{&\hbox{maximize\ \ }#1\cr &\hbox{subject to\ \
}#2\cr}\ifnum 0=#3{}\else\eqno(#3)\fi$$}        
     
\def\aligntwo#1#2#3#4#5{$$\eqalign{#1&#2\cr #3&#4\cr}
\ifnum 0=#5{}\else\eqno(#5)\fi$$}
\def\alignthree#1#2#3#4#5#6#7{$$\eqalign{#1&#2\cr #3&#4\cr #5&#6\cr}
\ifnum 0=#7{}\else\eqno(#7)\fi$$}


\def\eqnum{\eqno{\hbox{(\the\sectnum.\the\eqnumber)}\global\advance\eqnumber
by1}}

\def\eqnu{\eqno{\hbox{(\the\sectnum.\the\eqnumber)}\global\advance\eqnumber
by1}}

\newcount\examplnumber
\def\examplnum{\global\advance\examplnumber by1}

\newcount\figrnumber
\def\figrnum{\global\advance\figrnumber by1}

\newcount\propnumber
\def\propnum{\global\advance\propnumber by1}

\newcount\defnumber
\def\defnum{\global\advance\defnumber by1}

\newcount\lemmanumber
\def\lemmanum{\global\advance\lemmanumber by1}

\newcount\assumptionnumber
\def\assumptionnum{\global\advance\assumptionnumber by1}

\newcount\conditionnumber
\def\conditionnum{\global\advance\conditionnumber by1}

\def\exampl{\the\sectnum.\the\examplnumber}
\def\figr{\the\sectnum.\the\figrnumber}
\def\propn{\the\sectnum.\the\propnumber}
\def\defn{\the\sectnum.\the\defnumber}
\def\lemman{\the\sectnum.\the\lemmanumber}
\def\assumptionn{\the\sectnum.\the\assumptionnumber}
\def\condn{\the\sectnum.\the\conditionnumber}

\def\section#1{\goodbreak\vskip 3pc plus 6pt minus 3pt\leftskip=-2pc
   \global\advance\sectnum by 1\eqnumber=1
\global\examplnumber=1\figrnumber=1\propnumber=1\defnumber=1\lemmanumber=1\assumptionnumber=1 \conditionnumber =1%
   \line{\hfuzz=1pc{\hbox to 3pc{\bf 
   \vtop{\hfuzz=1pc\hsize=38pc\hyphenpenalty=10000\noindent\uppercase{\the\sectnum.\quad #1}}\hss}}
			\hfill}
			\leftskip=0pc\nobreak\tenf
			\vskip 1pc plus 4pt minus 2pt\noindent\ignorespaces}



\def\sect#1{\noindent\leftskip=-2pc\tenf
   \goodbreak\vskip 1pc plus 4pt minus 2pt
                \global\advance\subsectnum by 1\eqnumber=1
   \line{\hfuzz=1pc{\hbox to 3pc{\bf 
   \vtop{\hfuzz=1pc\hsize=38pc\hyphenpenalty=10000\noindent\uppercase{{\bf #1}}}\hss}}
                        \hfill}
   \leftskip=0pc\nobreak\tenf
                        \vskip 1pc plus 4pt minus 2pt\nobreak\noindent\ignorespaces}

\def\subsection#1{\noindent\leftskip=0pc\tenf
   \goodbreak\vskip 1pc plus 4pt minus 2pt
   \line{\hfuzz=1pc{\hbox to 3pc{\bf 
   \vtop{\hfuzz=1pc\hsize=38pc\hyphenpenalty=10000\noindent{\bf #1}}\hss}}
                        \hfill}
   \leftskip=0pc\nobreak\tenf
                        \vskip 1pc plus 4pt minus 2pt\nobreak\noindent\ignorespaces}
\def\subsubsection#1{\goodbreak\vskip 1pc plus 4pt minus 2pt
   \hfuzz=3pc\leftskip=0pc\noindent\tenit #1 \nobreak\tenf\vskip 6pt plus 1pt
                                minus 1pt\nobreak\ignorespaces\leftskip=0pc}
%

\def\beginexample#1{\noindent\goodbreak\vskip 6pt plus 1pt minus 1pt
\noindent
  \hbox {\bf Example #1\hss}
  \nobreak\vskip 4pt plus 1pt minus 1pt \nobreak\noindent\ninef
  \global\advance
                \leftskip by\parindent\pn}
\def\endexample{\vskip 12pt\tenf\par
  \global\advance\leftskip by -\parindent
  }

\def\beginexercise#1{\noindent\goodbreak\vskip 6pt plus 1pt minus 1pt \noindent\global\normalbaselineskip=12pt
  \hbox {\bf Exercise #1\hss}
  \nobreak\vskip 4pt plus 1pt minus 1pt 
  \nobreak\noindent\ninef\global\advance\leftskip
                        by\parindent\pn}
\def\endexercise{\vskip 12pt\tenf\par
  \global\advance\leftskip by -\parindent
  }

\def\beginsection#1{\noindent\goodbreak\vskip 6pt plus 1pt minus 1pt \noindent\global\normalbaselineskip=12pt
  \hbox {\it #1\hss}
  \vskip 0.1pt plus 1pt minus 1pt \nobreak\noindent\ninef\global\advance
                \leftskip by\parindent\noindent\pn}
\def\endsection{\vskip 12pt\tenf\par
  \global\advance\leftskip by -\parindent
}

%


\def\proposition#1{\smskip\pn{\bf Proposition #1}\quad}
\def\proof{\smskip\pn{\bf Proof:}\quad}

 \def\qed{\quad{\bf
Q.E.D.} \par\bigskip}
\def\ref{\smskip\pn}

\def\chapter#1#2{{\bf \centerline{\helbigbig
{#1}}}\bigskip\bigskip{\bf \centerline{\helbigbig
{#2}}}\bigskip\bigskip} 



\def\longpapertitle#1#2#3{{\bf \centerline{\helbigb
{#1}}}\bigskip{\bf \centerline{\helbigb
{#2}}}\bigskip\bigskip{\centerline{
by}}\bigskip{\bf \centerline{
{#3}}}\bigskip\bigskip} 


\def\nitem#1{\smskip\item{#1}}

\newcount\alphanum
\newcount\romnum

\def\alphaenumerate{\ifcase\alphanum \or (a)\or (b)\or (c)\or (d)\or (e)\or
(f)\or (g)\or (h)\or (i)\or (j)\or (k)\fi}
\def\romenumerate{\ifcase\romnum \or (i)\or (ii)\or (iii)\or (iv)\or (v)\or
(vi)\or (vii)\or (viii)\or (ix)\or (x)\or (xi)\fi}

\def\alist{\begingroup\vskip10pt\alphanum=1
\parskip=2pt\parindent=0pt \leftskip=3pc
\everypar{\llap{{\rm\alphaenumerate\hskip1em}}\advance\alphanum by1}}

\def\nolist{\begingroup\vskip10pt\alphanum=0
\parskip=2pt\parindent=0pt \leftskip=3pc
\everypar{\llap{\global\advance\alphanum by1(\the\alphanum)\hskip1em}}}

\def\romlist{\begingroup\vskip10pt\romnum=1
\parskip=2pt\parindent=0pt \leftskip=5pc
\everypar{\llap{{\rm\romenumerate\hskip1em}}\advance\romnum by1}}



\long\def\fig#1#2#3{\vbox{\vskip1pc\vskip#1
\prevdepth=12pt \baselineskip=12pt
\vskip1pc
\hbox to\hsize{\hfill\vtop{\hsize=25pc\noindent{\eightbf Figure #2\ }
{\eightpoint#3}}\hfill}}}

\long\def\widefig#1#2#3{\vbox{\vskip1pc\vskip#1
\prevdepth=12pt \baselineskip=12pt
\vskip1pc
\hbox to\hsize{\hfill\vtop{\hsize=28pc\noindent{\eightbf Figure #2\ }
{\eightpoint#3}}\hfill}}}

\long\def\table#1#2{\vbox{\vskip0.5pc
\prevdepth=12pt \baselineskip=12pt
\hbox to\hsize{\hfill\vtop{\hsize=25pc\noindent{\eightbf Table #1\ }
{\eightpoint#2}}\hfill}}}

 
\def\rightheadline#1{\headline{\tenrm\hfil #1}}


\long\def\leftfig#1#2{\vbox{\smskip\hsize=220pt
\vtop{{\noindent {\bf #1}}}
\smskip
\noindent
\vbox{{\noindent #2}}
}}

\long\def\rightfig#1#2#3{\vbox{\smskip\vskip#1
\prevdepth=12pt \baselineskip=12pt
\hsize=210pt
\smskip
\vbox{\noindent{\eightbold #2}
\hskip1em{\eightpoint#3}}
}}

\long\def\concept#1#2#3#4#5{\bigskip\hrule
\vbox{\hbox{\leftfig{#1}{#2} \hskip3em
\rightfig{#3}{#4}{#5}} \smskip}
\hrule\bigskip}


\long\def\bconcept#1#2#3#4#5#6#7{
\vbox{
\hbox to \hsize{\vtop{\par #1}}
\concept{#2}{#3}{#4}{#5}{#6}
\hbox to \hsize{\vtop{\par #7}}
\smskip}
}




\def\boxit#1{\vbox{\hrule\hbox{\vrule\kern3pt
                                \vbox{\kern3pt#1\kern3pt}\kern3pt\vrule}\hrule}}
\def\centerboxit#1{$$\vbox{\hrule\hbox{\vrule\kern3pt
                                \vbox{\kern3pt#1\kern3pt}\kern3pt\vrule}\hrule}$$}

\long\def\boxtext#1#2{$$\boxit{\vbox{\hsize #1\noindent\strut #2\strut}}$$}

%
%
%

\def\picture #1 by #2 (#3){
  \vbox to #2{
    \hrule width #1 height 0pt depth 0pt
    \vfill
    \special{picture #3} 
    }
  }

\def\scaledpicture #1 by #2 (#3 scaled #4){{
  \dimen0=#1 \dimen1=#2
  \divide\dimen0 by 1000 \multiply\dimen0 by #4
  \divide\dimen1 by 1000 \multiply\dimen1 by #4
  \picture \dimen0 by \dimen1 (#3 scaled #4)}
  }

%
%

\long\def\captfig#1#2#3#4#5{\vbox{\vskip1pc
\hbox to\hsize{\hfill{\picture #1 by #2 (#3)}\hfill}
\prevdepth=9pt \baselineskip=9pt
\vskip1pc
\hbox to\hsize{\hfill\vtop{\hsize=24pc\noindent{\eightbold Figure #4}
\hskip1em{\eightpoint#5}}\hfill}}}

%
%
%

\def\illustration #1 by #2 (#3){
  \vskip#2\hskip#1\special{illustration #3} 
    }

\def\scaledillustration #1 by #2 (#3 scaled #4){{
  \dimen0=#1 \dimen1=#2
  \divide\dimen0 by 1000 \multiply\dimen0 by #4
  \divide\dimen1 by 1000 \multiply\dimen1 by #4
  \illustration \dimen0 by \dimen1 (#3 scaled #4)}
  }


\newbox\graybox
\newdimen\xgrayspace
\newdimen\ygrayspace
%
%
%
%
%
%
%
%
%

\def\Textshade#1#2#3#4#5#6{%
    \xgrayspace=#4pt%
    \ygrayspace=#4pt%
    \def\grayshade{#3}%
    \def\linewidth{#5}%
    \def\theradius{#6}%
    \setbox\graybox=\hbox{\surroundboxa{#2}}%
    \hbox{%
    \hbox to 0pt{%
    \PScommands
}%
    \box\graybox}}%
%
%
\long%

\long%
\def\Parashade#1#2#3#4#5#6#7{%
    \xgrayspace=#4pt%
    \ygrayspace=#4pt%
    \def\grayshade{#3}%
    \def\linewidth{#5}%
    \def\theradius{#6}%
    \def\thevskip{#7pt}%
    \setbox\graybox=\hbox{\surroundboxb{#2}}%
    \vskip\thevskip%
    \hbox{%
    \hbox to 0pt{%
    \PScommands
}%
     \box\graybox}%
     \vskip\thevskip%
}%
%
%
%
\long\def\surroundboxa#1{\leavevmode\hbox{\vtop{%
\vbox{\kern\ygrayspace%
\hbox{\kern\xgrayspace#1%
      \kern\xgrayspace}}\kern\ygrayspace}}}
%
%
\long\def\surroundboxb#1{\leavevmode\hbox{\vtop{%
\vbox{\kern\ygrayspace%
\hbox{\kern\xgrayspace\vbox{\advance\hsize-2\xgrayspace#1}%
      \kern\xgrayspace}}\kern\ygrayspace}}}
%
%
%
\long\def\PScommands{%
\special{rawpostscript
/sharpbox{%
           newpath
           xmin ymin moveto
           xmin ymax lineto
           xmax ymax lineto
           xmax ymin lineto
           xmin ymin lineto
           closepath 
          }bind def
}%
\special{rawpostscript
/sharpboxnb{%
           newpath
           xmin ymin moveto
           xmin ymax lineto
           xmax ymax lineto
           xmax ymin lineto
          }bind def
}%
\special{rawpostscript
/sharpboxnt{%
           newpath
           xmin ymax moveto
           xmin ymin lineto
           xmax ymin lineto
           xmax ymax lineto
          }bind def
}%
\special{rawpostscript
/roundbox{%
           newpath
           xmin radius add ymin moveto
           xmax ymin xmax ymax radius arcto
           xmax ymax xmin ymax radius arcto
           xmin ymax xmin ymin radius arcto
           xmin ymin xmax ymin radius arcto 16 {pop} repeat
           closepath
          }bind def
}%
\special{rawpostscript
/sharpcorners{%
               sharpbox gsave grayshade setgray fill grestore 
               linewidth setlinewidth stroke
              }bind def
}%
\special{rawpostscript
/sharpcornersnt{%
               sharpboxnt gsave grayshade setgray fill grestore 
               linewidth setlinewidth stroke
              }bind def
}%
\special{rawpostscript
/sharpcornersnb{%
               sharpboxnb gsave grayshade setgray fill grestore 
               linewidth setlinewidth stroke
              }bind def
}%
\special{rawpostscript
/roundcorners{%
               roundbox gsave grayshade setgray fill grestore 
               linewidth setlinewidth stroke
              }bind def
}%
\special{rawpostscript
/plainbox{%
           sharpbox grayshade setgray fill 
          }bind def
}%
%
\special{rawpostscript
/roundnoframe{%
               roundbox grayshade setgray fill 
              }bind def
}%
\special{rawpostscript
/sharpnoframe{%
               sharpbox grayshade setgray fill 
              }bind def
}%
}%
%
%

\def\pshade#1{\Parashade{sharpcorners}{#1}{0.95}{10}{0.5}{10}{10}}


\def\boxit#1{\vbox{\hrule\hbox{\vrule\kern3pt
                                \vbox{\kern3pt#1\kern3pt}\kern3pt\vrule}\hrule}}

\def\boxitnb#1{\vbox{\hrule\hbox{\vrule\kern3pt
                                \vbox{\kern3pt#1\kern3pt}\kern3pt\vrule}}}

\def\boxitnt#1{\vbox{\hbox{\vrule\kern3pt
                                \vbox{\kern3pt#1\kern3pt}\kern3pt\vrule}\hrule}}

\long\def\boxtext#1#2{$$\boxit{\vbox{\hsize #1\noindent\strut #2\strut}}$$}



\def\texshopbox#1{\boxtext{462pt}{\vskip-1.5pc\pshade{\vskip-1.0pc#1\vskip-2.0pc}}}


%
%
%
%
%
%
%
%
\font\helbigbig=cmr10 scaled 2500%
\font\helbigb=cmbx10 scaled 1500%
\font\eightbold=cmbx8%

\def\tenf{\hel}%
\def\tenit{\heli}%
\def\ninef{\ninehel}%
\def\nineit{\nineheli}%
%
%


\font\tenrm=cmr10%
\font\teni=cmmi10%
\font\tensy=cmsy10%
\font\tenbf=cmbx10%
\font\tentt=cmtt10%
\font\tenit=cmti10%
\font\tensl=cmsl10%

\def\tenpoint{\def\rm{\fam0\tenrm}%
\textfont0=\tenrm%
\textfont1=\teni%
\textfont2=\tensy%
\textfont\itfam=\tenit%
\textfont\slfam=\tensl%
\textfont\ttfam=\tentt%
\textfont\bffam=\tenbf%
\scriptfont0=\sevenrm%
\scriptfont1=\seveni%
\scriptfont2=\sevensy%
\scriptscriptfont0=\sixrm%
\scriptscriptfont1=\sixi%
\scriptscriptfont2=\sixsy%
\def\it{\fam\itfam\tenit}%
\def\tt{\fam\ttfam\tentt}%
\def\sl{\fam\slfam\tensl}%
\scriptfont\bffam=\sevenbf%
\scriptscriptfont\bffam=\sixbf%
\def\bf{\fam\bffam\tenbf}%
\normalbaselineskip=18pt%
\normalbaselines\rm}%

\font\ninerm=cmr9%
\font\ninebf=cmbx9%
\font\nineit=cmti9%
\font\ninesy=cmsy9%
\font\ninei=cmmi9%
\font\ninett=cmtt9%
\font\ninesl=cmsl9%

\def\ninepoint{\def\rm{\fam0\ninerm}%
\textfont0=\ninerm%
\textfont1=\ninei%
\textfont2=\ninesy%
\textfont\itfam=\nineit%
\textfont\slfam=\ninesl%
\textfont\ttfam=\ninett%
\textfont\bffam=\ninebf%
\scriptfont0=\sixrm%
\scriptfont1=\sixi%
\scriptfont2=\sixsy%
\def\it{\fam\itfam\nineit}%
\def\tt{\fam\ttfam\ninett}%
\def\sl{\fam\slfam\ninesl}%
\scriptfont\bffam=\sixbf%
\scriptscriptfont\bffam=\fivebf%
\def\bf{\fam\bffam\ninebf}%
\normalbaselineskip=16pt%
\normalbaselines\rm}%

\font\eightrm=cmr8%
\font\eighti=cmmi8%
\font\eightsy=cmsy8%
\font\eightbf=cmbx8%
\font\eighttt=cmtt8%
\font\eightit=cmti8%
\font\eightsl=cmsl8%

\def\eightpoint{\def\rm{\fam0\eightrm}%
\textfont0=\eightrm%
\textfont1=\eighti%
\textfont2=\eightsy%
\textfont\itfam=\eightit%
\textfont\slfam=\eightsl%
\textfont\ttfam=\eighttt%
\textfont\bffam=\eightbf%
\scriptfont0=\sixrm%
\scriptfont1=\sixi%
\scriptfont2=\sixsy%
\scriptscriptfont0=\fiverm%
\scriptscriptfont1=\fivei%
\scriptscriptfont2=\fivesy%
\def\it{\fam\itfam\eightit}%
\def\tt{\fam\ttfam\eighttt}%
\def\sl{\fam\slfam\eightsl}%
\scriptscriptfont\bffam=\fivebf%
\def\bf{\fam\bffam\eightbf}%
\normalbaselineskip=14pt%
\normalbaselines\rm}%

\font\sevenrm=cmr7%
\font\seveni=cmmi7%
\font\sevensy=cmsy7%
\font\sevenbf=cmbx7%

\def\sevenpoint{%
   \def\rm{\sevenrm}\def\bf{\sevenbf}%
   \def\smc{\sevensmc}\baselineskip=12pt\rm}%

\font\sixrm=cmr6%
\font\sixi=cmmi6%
\font\sixsy=cmsy6%
\font\sixbf=cmbx6%

\fontdimen13\tensy=2.6pt%
\fontdimen14\tensy=2.6pt%
\fontdimen15\tensy=2.6pt%
\fontdimen16\tensy=1.2pt%
\fontdimen17\tensy=1.2pt%
\fontdimen18\tensy=1.2pt%

\def\tenf{\tenpoint}%
\def\ninef{\ninepoint}%
%




\def\section#1{\goodbreak\vskip 3pc plus 6pt minus 3pt\leftskip=-2pc
   \global\advance\sectnum by 1\eqnumber=1\subsectnum=0%
\global\examplnumber=1\figrnumber=1\propnumber=1\defnumber=1\lemmanumber=1\assumptionnumber=1 \conditionnumber =1%
   \line{\hfuzz=1pc{\hbox to 3pc{\bf 
   \vtop{\hfuzz=1pc\hsize=38pc\hyphenpenalty=10000\noindent\uppercase{\the\sectnum.\quad #1}}\hss}}
			\hfill}
			\leftskip=0pc\nobreak\tenf
			\vskip 1pc plus 4pt minus 2pt\noindent\ignorespaces}
\def\subsection#1{\noindent\leftskip=0pc\tenf
   \goodbreak\vskip 1pc plus 4pt minus 2pt
               \global\advance\subsectnum by 1
   \line{\hfuzz=1pc{\hbox to 3pc{\bf \the\sectnum.\the\subsectnum.
   \vtop{\hfuzz=1pc\hsize=38pc\hyphenpenalty=10000\noindent{\bf #1}}\hss}}
                        \hfill}
   \leftskip=0pc\nobreak\tenf
                        \vskip 1pc plus 4pt minus 2pt\nobreak\noindent\ignorespaces}



\def\texshopbox#1{\boxtext{462pt}{\vskip-1.5pc\pshade{\vskip-1.0pc#1\vskip-2.0pc}}}


\input miniltx

\ifx\pdfoutput\undefined
  \def\Gin@driver{dvips.def} 
\else
  \def\Gin@driver{pdftex.def} 
\fi

\input graphicx.sty
\resetatcatcode

\long\def\fig#1#2#3{\vbox{\vskip1pc\vskip#1
\prevdepth=12pt \baselineskip=12pt
\vskip1pc
\hbox to\hsize{\hfill\vtop{\hsize=30pc\noindent{\eightbf Figure #2\ }
{\eightpoint#3}}\hfill}}}

\def\show#1{}

\rightheadline{\botmark}

\pageno=1

\def\longpapertitle#1#2#3{{\bf \centerline{\helbigb
{#1}}}\medskip{\bf \centeline{\helbigb
{#2}}}\bigskip{\bf \centerline{
{#3}}}\bigskip}

\vskip-3pc

\def\xstar{X^{\raise0.04pt\hbox{\sevenpoint *}} }

\def\jstar{J^{\raise0.04pt\hbox{\sevenpoint *}} }
\def\qstar{Q^{\raise0.04pt\hbox{\sevenpoint *}} }

\rightheadline{\botmark}

\pageno=1

\rightheadline{\botmark}

\pageno=1

\rightheadline{\botmark}

\pn {\bf September 2019 (Revised April 2020)}
\bigskip \bigskip\bigskip

\bigskip

\def\longpapertitle#1#2#3{{\bf \centerline{\helbigb
{#1}}}\medskip{\bf \centerline{\helbigb
{#2}}}\bigskip{\bf \centerline{
{#3}}}\bigskip}

\vskip-3pc

\longpapertitle{Multiagent Rollout Algorithms and}{Reinforcement Learning}{ {Dimitri Bertsekas\footnote{\dag}{\ninepoint McAfee Professor of Engineering, MIT, Cambridge, MA, and Fulton Professor of Computational Decision Making, ASU, Tempe, AZ.}}}



\centerline{\bf Abstract}

\smskip
\pn We consider finite and infinite horizon dynamic programming problems, where the control at each stage consists of several distinct decisions, each one made by one of several agents. We introduce an approach, whereby at every stage, each agent's decision is made by executing a local rollout algorithm that uses a base policy, together with some coordinating  information from the other agents. The amount of local computation required at every stage by each agent is independent of the number of agents, while the amount of total computation (over all agents) grows linearly with the number of agents. By contrast, with the standard rollout algorithm, the amount of total computation grows exponentially with the number of agents. Despite the drastic reduction in required computation, we show that our algorithm has the fundamental cost improvement property of rollout: an improved performance relative to the base policy. We also discuss possibilities to improve further the method's computational efficiency through limited agent coordination and parallelization of the agents' computations. Finally, we explore related approximate policy iteration algorithms for infinite horizon problems, and we prove that the cost improvement property steers the algorithm towards convergence to an agent-by-agent optimal policy.

\vskip-1pc

\section{Multiagent Problem Formulation - Finite Horizon Problems}

\xdef\figfinitehorizon{\figr}\figrnum\show{myfigure}

\pn We consider a standard form of an $N$-stage dynamic programming (DP) problem (see [Ber17], [Ber19]), which involves the discrete-time dynamic system
$$x_{k+1}=f_k(x_k,u_k,w_k),\qquad k=0,1,\ldots,N-1,\xdef\systemeq{\lab}\eqnum\show{oneo}$$ 
where
$x_k$ is an element of some (possibly infinite) state space, the control $u_k$ is an  element of some finite control space, and $w_k$ is  a  random disturbance, which is characterized by a probability distribution $P_k
(\cdot\mid x_k,u_k)$ that may depend explicitly on $x_k$ and $u_k$, but not on 
values of prior disturbances $w_{k-1}, \ldots,w_0$. The control $u_k$ is constrained to take values in a given subset $U_k(x_k)$, which depends on the current state $x_k$. 
The cost of the $k$th stage is denoted by $g_k(x_k,u_k,w_k)$; see Fig.\ \figfinitehorizon. 

\topinsert
\centerline{\hskip0pc\includegraphics[width=5.8in]{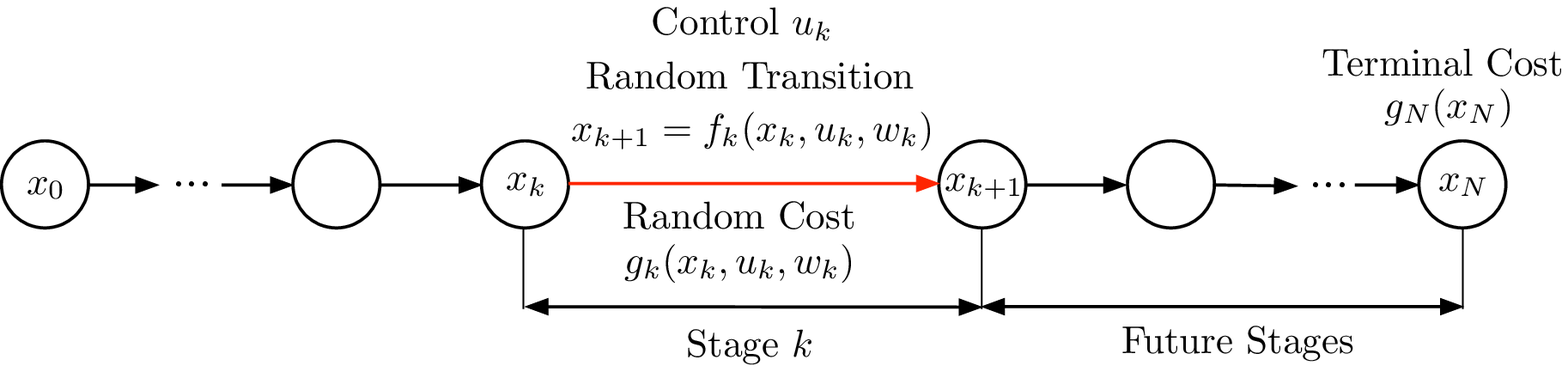}}
\vskip0pc
\hskip-3pc\fig{0pc}{\figfinitehorizon} {Illustration of the $N$-stage stochastic optimal control problem. Starting from state $x_k$, the next state under control $u_k$ is generated according to a system equation
$$x_{k+1}=f_k(x_k,u_k,w_k),$$
where $w_k$ is the random disturbance, and a random stage cost $g_k(x_k,u_k,w_k)$ is incurred.}
\endinsert

We consider policies of the form
$$\p=\{\m_0,\ldots,\m_{N-1}\},$$
where $\m_k$ maps states $x_k$ into controls $u_k=\m_k(x_k)$, and satisfies a control constraint of the form $\m_k(x_k)\in U_k(x_k)$ for all $x_k$. 
Given an initial state $x_0$ and a policy 
$\p=\{\m_0,\ldots,\m_{N-1}\}$, the expected cost of $\p$ starting at $x_0$ is
$$J_{\p}(x_0)=E\left\{g_{N}(x_N)+\sum_{k=0}^{N-1}g_k\big(x_k,\m_k(x_k),w_k\big)\right\},$$
where the expected value operation $E\{\cdot\}$ is over all the random variables $w_k$ and $x_k$. The optimal cost is the function $\jstar$ of the initial state $x_0$, defined by 
$$\jstar(x_0)=\min_{\p\in \P}J_{\p}(x_0),$$
where $\P$ is the set of all policies, while an optimal policy
$\p^*$ is one that attains the minimal cost for every initial state; i.e., 
$$J_{\p^*}(x_0)=\min_{\p\in \P}J_{\p}(x_0).$$
Since $\jstar$ and $\p^*$ are  typically hard to obtain by exact DP, we consider reinforcement learning (RL) algorithms for suboptimal solution, and focus on rollout, which we describe next.  

\subsection{The Standard Rollout Algorithm}
\vskip-0.5pc

\pn The aim of rollout is policy improvement. In particular, given a policy $\p=\{\m_0,\ldots,\m_{N-1}\},$ called {\it base policy\/}, with cost-to-go from state $x_k$ at stage $k$ denoted by $J_{k,\p}(x_k)$, $k=0,\ldots,N$, we wish to use rollout to obtain an improved policy, i.e., one that achieves cost that is at most $J_{k,\p}(x_k)$ starting from each $x_k$. The standard rollout algorithm provides on-line control of the system as follows (see the textbooks [BeT96], [Ber17], [Ber19]):

\texshopbox{
\pn{\bf Standard One-Step Lookahead Rollout Algorithm}:
\pn Start with the initial state $x_0$, and proceed forward generating a trajectory 
$$\{x_0,\tl u_0,x_1,\tl u_1,\ldots,x_{N-1},\tl u_{N-1},x_N\}$$
according to the system equation \systemeq, by applying at each state $x_k$ a control $\tl u_k$ selected by the one-step lookahead minimization
$$\tl u_k\in\arg\min_{u_k\in U_k(x_k)} E\Big\{g_k(x_k,u_k,w_k)+J_{{k+1},\p}\big(f_k(x_k,u_k,w_k)\big)\Big\}.\xdef\eqone{\lab}\eqnum\show{oneo}$$
}

The one-step minimization \eqone, which uses $J_{{k+1},\p}$ in place of the optimal cost-to-go function, defines a policy $\tl \p=\{\tl \m_0,\ldots,\tl \m_{N-1}\}$, where for all $x_k$ and $k$, $\tl \m_k(x_k)$ is equal to the control $\tl u_k$ obtained from Eq.\ \eqone. This policy is referred to as the {\it rollout policy\/}. The fundamental cost improvement result here is that the rollout policy improves over the base policy in the sense that
$$J_{k,\tl \p}(x_k)\le J_{k,\p}(x_k),\qquad \forall\ x_k, k,\xdef\eqtwo{\lab}\eqnum\show{oneo}$$
where $J_{k,\tl\p}(x_k)$, $k=0,\ldots,N$, is the cost-to-go of the rollout policy starting from state $x_k$  ([Ber17], Section 6.4, or [Ber19], Section 2.4.2).

The expected value in Eq.\ \eqone\ is the Q-factor of the pair $(x_k,u_k)$ corresponding to the base policy:
$$Q_{k,\p}(x_k,u_k)=E\Big\{g_k(x_k,u_k,w_k)+J_{{k+1},\p}\big(f_k(x_k,u_k,w_k)\big)\Big\}.$$
 In the ``standard" implementation of rollout, at each encountered state $x_k$, the Q-factor $Q_{k,\p}(x_k,u_k)$ is computed by some algorithm separately for each control $u_k\in U_k(x_k)$ (often by Monte Carlo simulation). Unfortunately, in the multiagent context to be discussed shortly, the number of controls in $U_k(x_k)$, and the attendant computation of Q-factors, grow rapidly with the number of agents, and can become very large. The purpose of this paper is to introduce a modified rollout algorithm for the multiagent case, which requires much less computation while maintaining the cost improvement property \eqtwo.
 
 \vskip-0.5pc

\subsection{The Multiagent Case}
\vskip-0.5pc

\pn Let us  assume a special structure of the control space, corresponding to a multiagent version of the problem.\footnote{\dag}{\ninepoint While we focus on multiagent problems, our methodology applies to any problem where the control $u_k$ consists of $m$ components, $u_k=(u_k^1,\ldots,u_k^m)$.} In particular, we assume that the control $u_k$ consists of $m$ components $u_k^1,\ldots,u_k^m$, 
$$u_k=(u_k^1,\ldots,u_k^m),$$
with the component $u_k^\ell$, $\ell=1,\ldots,m$, chosen by agent $\ell$ at stage $k$, from within a given set $U_k^\ell(x_k)$. Thus  the control constraint set is the Cartesian product\footnote{\dag}{\ninepoint The Cartesian product structure of the constraint set is adopted here for simplicity of exposition, particularly when arguing about computational complexity. The idea of trading off control space complexity and state space complexity (cf.\ Section 1.3), on which this paper rests, does not depend on a Cartesian product constraint structure. Of course when this structure is present, it simplifies the computations of the methods of this paper.}
$$U_k(x_k)=U_k^1(x_k)\times \cdots \times U_k^m(x_k).\xdef\eqthree{\lab}\eqnum\show{oneo}$$
Then the minimization \eqone\ involves as many as $s^m$ Q-factors, where $s$ is the maximum number of elements of the sets $U_k^i(x_k)$ [so that $s^m$ is an upper bound to the number of controls in $U_k(x_k)$, in view of its Cartesian product structure \eqthree]. Thus the computation required by the rollout algorithm is of order $O(s^m)$ per stage.

In this paper we propose an alternative rollout algorithm that  achieves the cost improvement property \eqtwo\ at much smaller computational cost, namely of order $O(sm)$ per stage. A key idea here is that the computational requirements of the rollout one-step minimization \eqone\ are proportional to the number of controls in the set $U_k(x_k)$ and are independent of the size of the state space. This motivates a reformulation of the problem, first suggested in the neuro-dynamic programming book [BeT96], Section 6.1.4, whereby control space complexity is traded off with state space complexity by ``unfolding" the control $u_k$ into its $m$ components, which are applied one {\it agent-at-a-time} rather than {\it all-agents-at-once\/}. We discuss this idea next within the multiagent context.

\subsection{Trading off Control Space Complexity with State Space 
Complexity}
\vskip-0.5pc

\pn We noted that a major issue in rollout is the  minimization over $u_k\in U_k(x_k)$ in Eq.\ \eqone, which may be very time-consuming when the size of the control constraint set is large. In particular, in the multiagent case where $u_k=(u_k^1,\ldots,u_k^m),$ the time to perform this minimization is typically exponential in $m$. In this case, we can reformulate the problem by breaking down the collective decision $u_k$ into $m$ individual component decisions, thereby reducing the complexity of the control space while
increasing the complexity of
the state space. The potential advantage is that the extra state space 
complexity does not affect the computational requirements of some RL algorithms, including rollout.

\xdef\figunfolded{\figr}\figrnum\show{myfigure}

To this end, we introduce a modified but equivalent problem, involving {\it one-agent-at-a-time control selection\/}. At the generic state $x_k$, we break down the control $u_k$ into the 
sequence
of the $m$ controls $u_k^1,u_k^2,\ldots,u_k^m$, and between $x_k$ and the next state $x_{k+1}=f_k(x_k,u_k,w_k)$, we introduce artificial 
intermediate ``states" $(x_k,u_k^1),(x_k,u_k^1,u_k^2),\ldots,(x_k,u_k^1,\ldots,u_k^{m-1})$, and corresponding 
transitions. The choice of the last control component $u_k^m$ 
at ``state" $(x_k,u_k^1,\ldots,u_k^{m-1})$ marks the transition to the next state $x_{k+1}=f_k(x_k,u_k,w_k)$ according to 
the system equation, while incurring cost $g_k(x_k,u_k,w_k)$; see Fig.\ \figunfolded.

\topinsert
\centerline{\hskip0pc\includegraphics[width=5.8in]{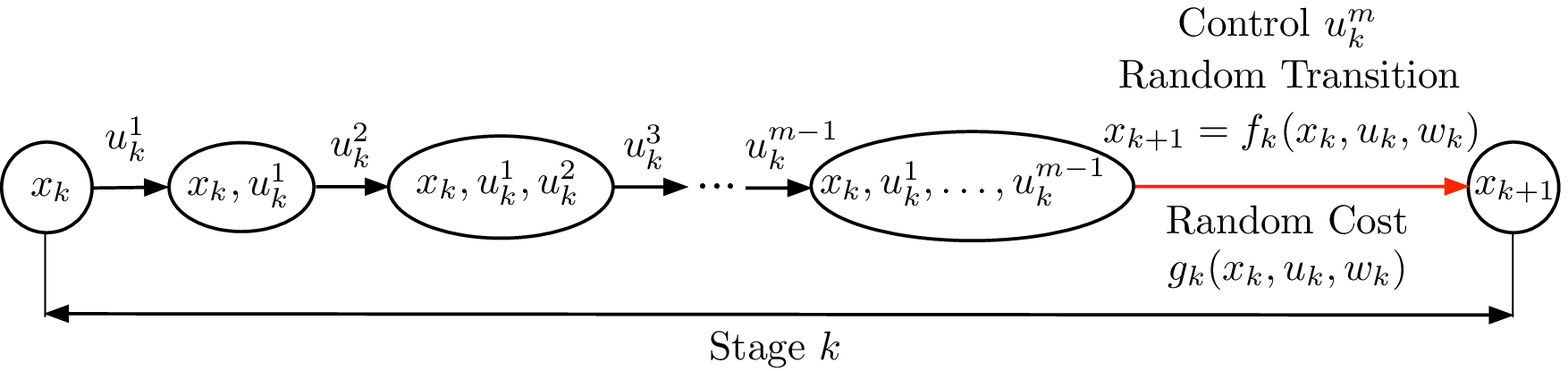}}\vskip0pc
\hskip-3pc\fig{-1.0pc}{\figunfolded} {Equivalent formulation of the $N$-stage stochastic optimal control problem for the case where the control $u_k$ consists of $m$ components $u_k^1,u_k^2,\ldots,u_k^m$:
$$u_k=(u_k^1,\ldots,u_k^m)\in U_k^1(x_k)\times \cdots \times U_k^m(x_k).$$
The figure depicts the $k$th stage transitions. Starting from state $x_k$, we generate the intermediate states $(x_k,u_k^1),(x_k,u_k^1,u_k^2),\ldots,(x_k,u_k^1,\ldots,u_k^{m-1})$, using the respective controls $u_k^1,\ldots, u_k^{m-1}$. The final control $u^m$ leads from $(x_k,u_k^1,\ldots,u_k^{m-1})$ to $x_{k+1}=f_k(x_k,u_k,w_k)$, and a random stage cost $g_k(x_k,u_k,w_k)$ is incurred.}
\endinsert

It is evident that this reformulated problem is equivalent to the original, since any control choice that is possible in one problem is also possible in the other problem, while the cost structure of the two problems is the same. In particular, every policy 
$$\p=\big\{(\m_k^1,\ldots,\m_k^m)\mid k=0,\ldots,N-1\big\}$$
of the original problem, including a base policy in the context of rollout, is admissible for the reformulated problem, and has the same cost function for the original as well as the reformulated problem.\footnote{\dag}{\ninepoint  It would superficially appear that the reformulated problem contains more policies than the original problem, because its form is more general: in the reformulated problem the policy at the $k$th stage applies the control components  
$$\m_k^1(x_k),\m_k^2(x_k,u_k^1),\ldots,\m_k^m(x_k,u_k^1,\ldots,u_k^{m-1}),$$
where $u_k^1=\m_k^1(x_k)$ and for $\ell=2,\ldots,m$, $u_k^\ell$ is defined sequentially as
$$u_k^\ell=\m_k^\ell(x_k,u_k^1,\ldots,u_k^{\ell-1}).$$
Still, however, this policy is equivalent to the policy of the original problem that applies  the control components  
$$\m_k^1(x_k),\hat\m_k^2(x_k),\ldots,\hat \m_k^m(x_k),$$
where for $\ell=2,\ldots,m$, $\hat\m_k^\ell(x_k)$ is defined sequentially as
$$\hat\m_k^\ell(x_k)=\m_k^\ell\big(x_k,\m_k^1(x_k),\hat\m_k^2(x_k),\ldots,
\hat\m_k^{\ell-1}(x_k)\big).$$}

The motivation for the reformulated problem is that the control space is simplified 
at the expense of
introducing $m-1$ additional layers of states, and corresponding $m-1$
cost-to-go functions
$J_k^1(x_k,u_k^1)$, $J_k^2(x_k,u_k^1,u_k^2), \ldots,J_k^{m-1}(x_k,u_k^1,\ldots,u_k^{m-1})$, in addition to $J_k(x_k)$.  On the other hand, the increase in size of the state 
space does not adversely affect the operation of rollout, since the Q-factor minimization \eqone\ is performed for just one state at each stage. 
Moreover, in a different context, the increase in size of the state space can be dealt with by
using function approximation, i.e., with the introduction of 
cost-to-go approximations
$$\tl J_k^1(x_k,u_k^1,r_k^1),\  \tl J_k^2(x_k,u_k^1,u_k^2,r_k^2), \ldots,\tl 
J_k^{m-1}(x_k,u_k^1,\ldots,u_k^{m-1},r_k^{m-1}),$$
in
addition to $\tl J_k(x_k,r_k)$, where $r_k,r_k^1,\ldots,r_k^{m-1}$ are parameters of corresponding approximation architectures (such as feature-based architectures and neural networks).

\vskip-1pc

\section{Multiagent Rollout}

\vskip-0.5pc

\pn Consider now the standard rollout algorithm applied to the reformulated problem shown in Fig.\ \figunfolded, with a given base policy $\p=\{\m_0,\ldots,\m_{N-1}\},$ which is also a policy of the original problem [so that $\m_k=(\m_k^1,\ldots,\m_k^m)$, with each $\m_k^\ell$, $\ell=1,\ldots,m$, being a function of just $x_k$]. The algorithm generates a rollout policy $\tl \p=\{\tl \m_0,\ldots,\tl \m_{N-1}\},$ where for each stage $k$, $\tl \m_k$ consists of $m$ components $\tl \m_k^\ell$, i.e., 
$\tl \m_k=(\tl \m_k^1,\ldots,\tl \m_k^m),$
and is obtained for all  $x_k$ according to 
{\ninepoint $$\tl \m_k^1(x_k)\in\arg\min_{u_k^1\in U_k^1(x_k)}E\Big\{g_k\big(x_k,u_k^1,\m_k^2(x_k),\ldots,\m_k^m(x_k),w_k\big)+J_{k+1,\p}\Big(f_k\big(x_k,u_k^1,\m_k^2(x_k),\ldots,\m_k^m(x_k),w_k\big)\Big)\Big\},$$
$$\tl \m_k^2(x_k)\in\arg\min_{u_k^2\in U_k^2(x_k)}E\Big\{g_k\big(x_k,\tl \m_k^1(x_k),u_k^2,\ldots,\m_k^m(x_k),w_k\big)+J_{k+1,\p}\Big(f_k\big(x_k,\tl \m_k^1(x_k),u_k^2,\ldots,\m_k^m(x_k),w_k\big)\Big)\Big\},$$
$$\cdots\qquad\qquad\cdots\qquad\qquad\cdots$$
$$\tl \m_k^m(x_k)\in\arg\min_{u_k^m\in U_k^m(x_k)}E\Big\{g_k\big(x_k,\tl \m_k^1(x_k),\tl \m_k^2(x_k),\ldots,u_k^m,w_k\big)+J_{k+1,\p}\Big(f_k\big(x_k,\tl \m_k^1(x_k),\tl \m_k^2(x_k),\ldots,u_k^m,w_k\big)\Big)\Big\}.\xdef\multiagentrollout{\lab}\eqnum\show{oneo}$$}

Thus, when applied on-line, at  $x_k$, {\it the algorithm generates the control $\tl \m_k(x_k)=\big(\tl \m_k^1(x_k),\ldots,\tl \m_k^m(x_k)\big)$ via a sequence of $m$ minimizations, once over each of the agent controls $u_k^1,\ldots,u_k^m$, with the past controls determined by the rollout policy, and the future controls determined by the base policy\/}; cf.\ Eq.\ \multiagentrollout.  Assuming a maximum of $s$ elements in the constraint sets $U_k^i(x_k)$, the computation required at each stage $k$ is of order $O(n)$ for each of the ``states" 
$x_k,(x_k,u_k^1),\ldots,(x_k,u_k^1,\ldots,u_k^{m-1}),$
 for a total of order $O(sm)$ computation. 

In the ``standard" implementation of the algorithm, at each $(x_k,u_k^1,\ldots,u_k^{\ell-1})$ with $\ell\le m$, and for each of the controls $u_k^\ell$, we generate by simulation a number of system trajectories up to stage $N$, with all future controls determined by the base policy. We average the costs of these trajectories, thereby obtaining the $Q$-factor corresponding to $(x_k,u_k^1,\ldots,u_k^{\ell-1},u_k^\ell)$. We then select the control $u_k^\ell$ that corresponds to the minimal $Q$-factor, with the controls $u_k^1,\ldots,u_k^{\ell-1}$ held fixed at the values computed earlier.

Prerequisite assumptions for the preceding algorithm to work in an on-line multiagent setting are:

\nitem{(a)} All agents have access to the current state $x_k$.
\nitem{(b)} There is an order in which agents compute and apply their local controls.
\nitem{(c)} There is intercommunication between agents, so agent $\ell$ knows the local controls  $u_k^1,\ldots,u_k^{\ell-1}$ computed by the predecessor agents $1,\ldots,\ell-1$ in the given order.  
\smskip

Note that the rollout policy \multiagentrollout, obtained from the reformulated problem is different from the rollout policy obtained from the original problem [cf.\ Eq.\ \eqone]. Generally, it is unclear how the two rollout policies perform relative to each other in terms of attained cost. 
On the other hand, both rollout policies perform no worse than the base policy, since the performance of the base policy is identical for both the reformulated problem and for the original problem. This is shown formally in the following proposition. 

\xdef\propcostimprove{\propn}\propnum\show{myproposition}

\texshopbox{\proposition{\propcostimprove:}Let $\p$ be a base policy and let $\tl\p$ be a corresponding rollout policy generated by the multiagent rollout algorithm \multiagentrollout. We have
$$J_{k,\tl\p}(x_k)\le J_{k,\p}(x_k),\qquad \hbox{for all $x_k$ and $k$.}\xdef\costimprstoch{\lab}\eqnum\show{oneo}$$
}

\proof We will show Eq.\ \costimprstoch\ by induction, and for simplicity, we will give the proof for the case of just two agents, i.e., $m=2$.
Clearly Eq.\ \costimprstoch\ holds for $k=N$, since $J_{N,\tl\p}= J_{N,\p}=g_N$. Assuming that it holds for index $k+1$, i.e., $J_{k+1,\tl\p}\le J_{k+1,\p}$, we have for all $x_k$,
$$\eqalign{J_{k,\tl\p}(x_k)&=E\Big\{g_k\big(x_k,\tl \m_k^1(x_k),\tl \m_k^2(x_k),w_k\big)+J_{k+1,\tl\p}\Big(f_k\big(x_k,\tl \m_k^1(x_k),\tl \m_k^2(x_k),w_k\big)\Big)\Big\}\cr
&\le E\Big\{g_k\big(x_k,\tl \m_k^1(x_k),\tl \m_k^2(x_k),w_k\big)+J_{k+1,\p}\Big(f_k\big(x_k,\tl \m_k^1(x_k),\tl \m_k^2(x_k),w_k\big)\Big)\Big\}\cr
&=\min_{u_k^2\in U_k^2(x_k)}E\Big\{g_k(x_k,\tl \m_k^1(x_k),u_k^2,w_k)+J_{k+1,\p}\Big(f_k\big(x_k,\tl \m_k^1(x_k),u_k^2,w_k\big)\Big)\Big\}\cr
&\le E\Big\{g_k\big(x_k,\tl \m_k^1(x_k),\m_k^2(x_k),w_k\big)+J_{k+1,\p}\Big(f_k\big(x_k,\tl \m_k^1(x_k),\m_k^2(x_k),w_k\big)\Big)\Big\}\cr
&=\min_{u_k^1\in U_k^1(x_k)}E\Big\{g_k(x_k,u_k^1,\m_k^2(x_k),w_k)+J_{k+1,\p}\Big(f_k\big(x_k,u_k^1,\m_k^2(x_k),w_k\big)\Big)\Big\}\cr
&\le E\Big\{g_k\big(x_k,\m_k^1(x_k),\m_k^2(x_k),w_k\big)+J_{k+1,\p}\Big(f_k\big(x_k,\m_k^1(x_k),\m_k^2(x_k),w_k\big)\Big)\Big\}\cr
&=J_{k,\p}(x_k),\cr}\xdef\inductionarg{\lab}\eqnum\show{oneo}$$
where in the preceding relation:
\nitem{(a)} The first equality is the DP equation for the rollout policy $\tl \p$.
\nitem{(b)} The first inequality holds by the induction hypothesis.
\nitem{(c)} The second equality holds by the definition of the rollout algorithm as it pertains to agent 2.
\nitem{(d)} The third equality holds by the definition of the rollout algorithm as it pertains to agent 1.
\nitem{(e)} The last equality is the DP equation for the base policy $\p$.
\smskip
\pn The induction proof of the cost improvement property \costimprstoch\ is thus complete for the case $m=2$. The proof for an arbitrary number of agents $m$ is entirely similar. \qed

Note the difference in the proof argument between the all-agents-at-once and one-agent-at-a-time rollout algorithms. In the former algorithm, the second equality in Eq.\ \inductionarg\ would be over both   $u_k^1\in U_k^1(x_k)$ and $u_k^2\in U_k^2(x_k)$, and the second inequality and third equality would be eliminated. Still the proof of the cost improvement property \costimprstoch\ goes through in both cases. Note also that if the base policy were optimal, Eq.\ \inductionarg\ would hold as an equality throughout for both  rollout algorithms, while the rollout policy $\tl \p$ would also be optimal. 

On the other hand, there is an important situation where the all-agents-at-once rollout algorithm can improve the base policy but the one-agent-at-a-time algorithm will not. This possibility may arise when the base policy is  ``agent-by-agent-optimal,"  i.e., each agent's control component is optimal, assuming that the control components of all other agents are kept fixed at some known values.\footnote{\dag}{\ninepoint  This is a concept that has received much attention in the theory of team optimization, where it is known as {\it person-by-person optimality\/}. It has been studied in the context of somewhat different problems, which involve imperfect state information that may not be shared by all the agents;  see Marschak [Mar55], Radner [Rad62], Witsenhausen [Wit71], Ho [Ho80]. For more  recent works, see Nayyar, Mahajan, and Teneketzis [NMT13], Nayyar and Teneketzis [NaT19],  Li et al.\ [LTZ19], the book by Zoppoli, Parisini, Baglietto, and Sanguineti [ZPB19], and the references quoted there.} Such a policy may not be optimal, except under special conditions. Thus if the base policy is agent-by-agent-optimal, multiagent rollout will be unable to improve strictly the cost function, even if this base policy is strictly suboptimal. However, we speculate that a situation where a base policy is agent-by-agent-optimal is unlikely to occur in rollout practice, since ordinarily a base policy must be reasonably simple, readily available, and easily simulated.

Let us also note that the qualitative difference between all-agents-at-once versus one-agent-at-a-time rollout is reminiscent of the context of value iteration (VI) algorithms, which involve minimization of a Bellman equation-like expression over the control constraint. In such algorithms one may choose between Gauss-Seidel methods, where the cost of a single state (and the control at that state) is updated at a time, while taking into account the results of earlier state cost computations, and  Jacobi methods, where the cost of all  states is updated at once. The tradeoff between Gauss-Seidel and Jacobi methods is well-known in the VI context: generally, Gauss-Seidel methods are faster, while Jacobi methods are also valid, as well as better suited for distributed asynchronous implementation; see [BeT89], [Ber12]. Our context in this paper is quite different, however, since we are considering updates of agent controls, and not cost updates at different states.

Let us provide an example that illustrates how the size of the control space may become intractable for even moderate values of the number of agents $m$.

\xdef\figspiderandfly{\figr}\figrnum\show{myfigure}

\xdef\examplespiderfly{\exampl}\examplnum\show{myexample}

\beginexample{\examplespiderfly\ (Spiders and Fly)}Here there are $m$ spiders and one fly moving on a 2-dimensional grid.  During each time period the fly moves to a some other position according to a given state-dependent probability distribution. The spiders, working as a team, aim to catch the fly at minimum cost. Each spider learns the current state  (the vector of spiders and fly locations) at the beginning of each time period, and either moves to a neighboring location or stays where it is. Thus each spider $i$ has as many as five choices at each time period (with each move possibly incurring a different location-dependent cost). The control vector is $u=(u^1,\ldots,u^m)$, where $u^i$ is the choice of the $i$th spider, so there are about $5^m$ possible values of $u$. However, if we view this as a multiagent problem, as per the reformulation of Fig.\ \figunfolded, the size of the control space is reduced to $\le 5$ moves per spider.

\topinsert
\centerline{\hskip0pc\includegraphics[width=3.2in]{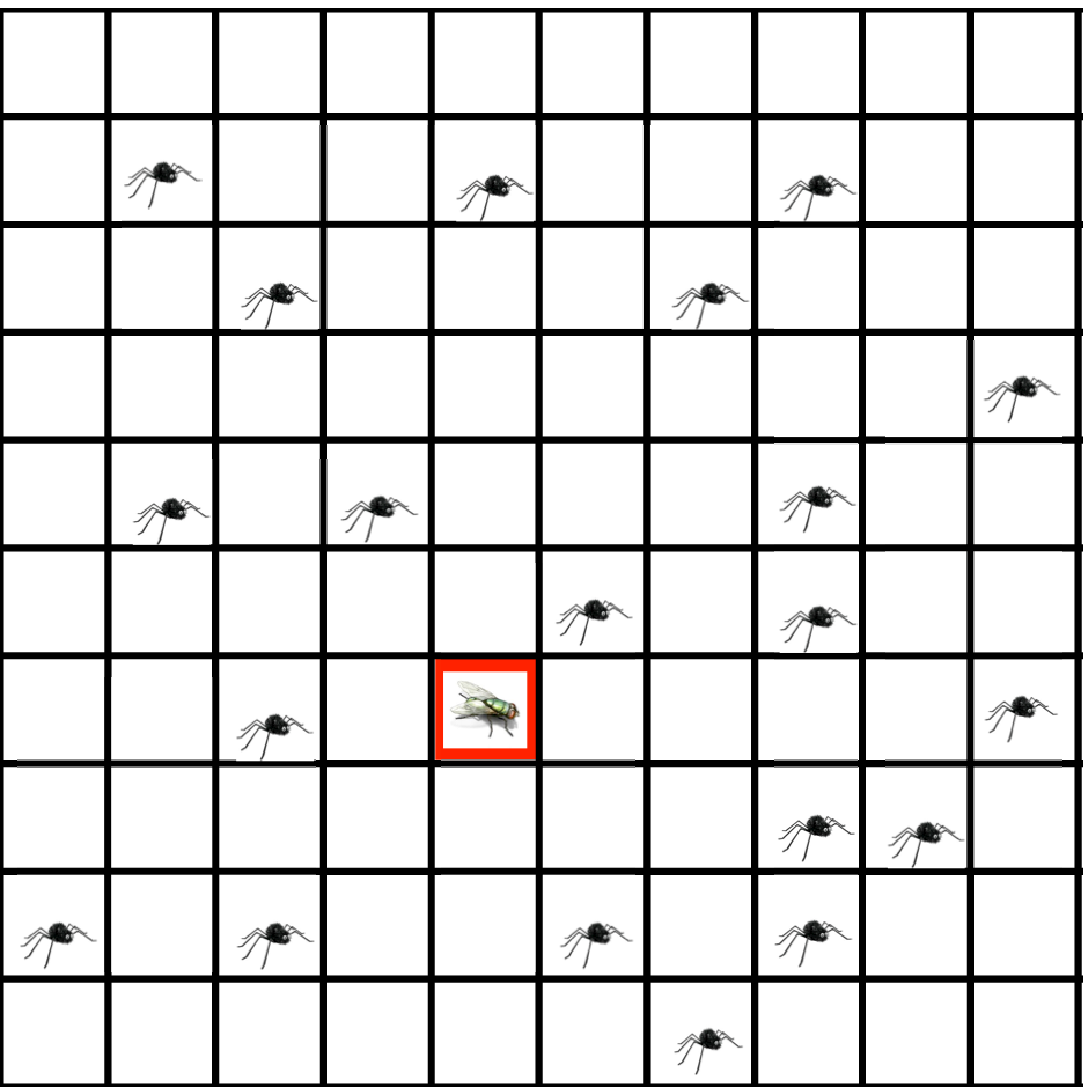}}
\vskip-1pc
\fig{0pc}{\figspiderandfly}{Illustration of the 2-dimensional spiders-and-fly problem. The state is the pair of distances between spiders and fly. At each time period, each spider moves to a neighboring location or stays where it is. The spiders make moves with perfect knowledge of the locations or each other and of the fly. The fly moves randomly, regardless of the position of the spiders.} 
\endinsert 

To apply multiagent rollout, we need a base policy. A simple possibility is to use the policy that directs each spider to move on the path of minimum distance to the current fly position. According to the multiagent rollout formalism, the spiders choose their moves in a given order, taking into account the current state,  and assuming that future moves will be chosen according to the base policy. This is a tractable computation, particularly if the rollout with the base policy is truncated after some stage, and the cost of the remaining stages is approximated  using a certainty equivalence approximation in order to reduce the cost of the Monte Carlo simulation.
The problem can be made more complicated by introducing terrain obstacles, travel costs, or multiple flies. 

Sample computations with this example indicate that the multiagent rollout algorithm of this section performs about as well as the standard rollout algorithm. Both algorithms perform much better than the base policy, and exhibit some ``intelligence" that the base policy does not possess. In particular, in the rollout algorithms the spiders attempt to ``encircle" the fly for faster capture, rather that moving straight towards the fly along a shortest path.  
\endexample

The following example is similar to the preceding one, but involves two flies and two spiders moving along a line, and admits an exact analytical solution. It illustrates how the multiagent rollout policy may exhibit intelligence and agent coordination that is totally lacking from the base policy. The behavior described in the example has been supported by computational experiments with larger two-dimensional problems of the type described in the preceding example.

\xdef\examplespidersandflies{\exampl}\examplnum\show{examplo}

\xdef\figspidersandflies{\figr}\figrnum\show{myfigure}

\beginexample{\examplespidersandflies\ (Spiders and Flies)}This is a spiders-and-flies problem that admits an analytical solution. There are two spiders and  two flies moving along integer locations on a straight line. For simplicity we will assume that the flies' positions are fixed at some integer locations, although the problem is qualitatively similar when the flies move randomly. The spiders have the option of moving either left or right by one unit; see Fig.\ \figspidersandflies. The objective is to minimize the time to capture both flies. The problem has essentially a finite horizon since the spiders can force the capture of the flies within a known number of steps.

\topinsert
\centerline{\hskip0pc\includegraphics[width=5.0in]{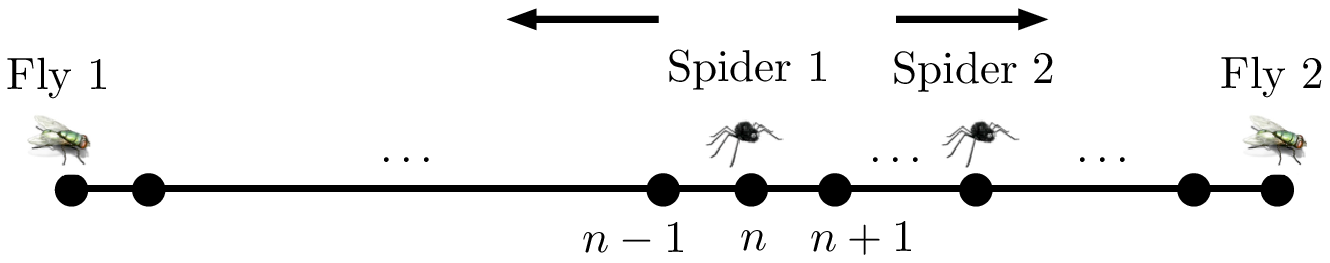}}\vskip0pc
\hskip-4.5pc
\fig{0pc}{\figspidersandflies}{Illustration of  the two-spiders and two-flies problem. The spiders move along integer points of a line. The two flies stay still at some integer locations. The optimal policy is to move the two spiders towards different flies, the ones that are initially closest to them. The base policy directs each spider to move one unit towards the nearest fly position.

Multiagent rollout with the given base policy starts with spider 1 at location $n$, and calculates the two Q-factors that correspond to moving to locations $n-1$ and $n+1$, assuming that the remaining moves of the two spiders will be made using the go-towards-the-nearest-fly base policy. The Q-factor of going to $n-1$ is smallest because it saves in unnecessary moves of spider 1 towards fly 2, so spider 1 will move towards fly 1. The trajectory generated by multiagent rollout is to move spiders 1 and 2 towards flies 1 and 2, respectively, then spider 2 first captures fly 2, and then spider 1 captures fly 1. Thus multiagent rollout generates the optimal policy.} 
\endinsert 

Here the optimal policy is to move the two spiders towards different flies, the ones that are initially closest to them (with ties broken arbitrarily). The minimal time to capture is the maximum of the two initial distances of the two optimal spider-fly pairings. 

Let us apply multiagent rollout with the base policy that directs each spider to move one unit towards the closest fly position (and in case of a tie, move towards the fly that lies to the right). The base policy is poor because it may unnecessarily move both spiders in the same direction, when in fact only one is needed to capture the fly.
This limitation is due to the lack of coordination between the spiders: each acts selfishly, ignoring the presence of the other. We will see that rollout  restores a significant degree of coordination between the spiders through an optimization that takes into account the long-term consequences of the spider moves.

According to the multiagent rollout mechanism, the spiders choose their moves one-at-a-time, optimizing over the two Q-factors corresponding to the right and left moves,  while assuming that future moves will be chosen according to the base policy. Let us consider a stage, where the two flies are alive while the spiders are at different locations as in  Fig.\ \figspidersandflies. Then the rollout algorithm will start with spider 1 and calculate two Q-factors corresponding to the right and left moves, while using the base heuristic to obtain the next move of spider 2, and the remaining moves of the two spiders. Depending on the values of the two Q-factors, spider 1 will move to the right or to the left, and it can be seen that it will choose to {\it move away from spider 2} even if doing so increases its distance to its closest fly {\it contrary to what the base heuristic will do\/}; see Fig.\ \figspidersandflies. Then spider 2 will act similarly and the process will continue. Intuitively, spider 1 moves away from spider 2 and fly 2, because it recognizes that spider 2 will capture earlier fly 2, so it might as well move towards the other fly. 

Thus {\it the multiagent  rollout algorithm induces implicit move coordination\/}, i.e., each spider moves in a way that takes into account future moves of the other spider. In fact it can be verified that the algorithm will produce an optimal sequence of moves starting from any initial state. It can also be seen that ordinary rollout (both flies move at once) will also produce an optimal move sequence. Moreover, the example admits a two-dimensional generalization, whereby the two spiders, starting from the same position, will separate under the rollout policy,  with each moving towards a different spider, while they will move in unison in the base policy whereby they move along the shortest path to the closest surviving fly. Again this will typically happen for both standard and  multiagent rollout.
\endexample

The preceding example illustrates how a poor base policy can produce a much better rollout policy, something that can be observed in many other problems. Intuitively, the key fact is that rollout is ``farsighted"  in the sense in can benefit from control calculations that reach far into future stages.

\vskip-1pc

\section{Rollout Variants for Finite Horizon Problems}

\vskip-0.5pc

\pn  It is worth noting a few variants of the rollout algorithm for the reformulated finite horizon problem of Fig.\ \figunfolded.
\nitem{(a)} Instead of selecting the agent controls in a fixed order, it is
possible to change the order at each stage $k$ (the preceding cost improvement proof goes through again by induction). In fact it is possible to optimize over multiple orders at the same stage, or to base the order selection on various features of the state $x$ (for instance in the case of Example  \examplespiderfly, with multiple spiders and flies, giving priority to the spiders ``closest" to some fly may make sense). 

\nitem{(b)} We can use at each stage $k$, a base policy $\{\m^1_k,\ldots,\m^m_k\}$ that selects controls that depend not just on the state $x_k$, but also on the preceding controls, i.e., consists of functions $\m_k^\ell$ of the form $\m_k^\ell(x_k,u^1_k,\ldots,u_k^{\ell-1})$. This can exploit intuition into the problem's structure, but from a theoretical viewpoint, it is not more general. The reason is that policies where control component selections $u_k^\ell$ depend on the previously selected control components $u^1_k,\ldots,u_k^{\ell-1}$ in addition to the current state $x_k$, can be equivalently represented by policies where the selection of $u_k^\ell$ depends on just $x_k$.

\nitem{(c)} The algorithm can be applied to a partial state information problem (POMDP), after it has been transformed to a perfect state information problem, using a belief state formulation, where the conditional probability distribution of the state given the available information plays the role of $x_k$ (note here that we have allowed the state space to be infinite, thereby making our methodology applicable to the POMDP/belief state formulation). 

\nitem{(d)} We may use rollout variants involving multistep lookahead, truncated rollout, and  terminal cost function approximation, in the manner described in the RL book [Ber19].  Of course, in such variants the cost improvement property need not hold strictly, but it holds within error bounds, some of which are given in [Ber19], Section 5.1, for the infinite horizon discounted problem case.
\smskip

We may also consider multiagent rollout algorithms that are asynchronous in the sense that the agents may compute their rollout controls in parallel or in some irregular order rather than in sequence, and they may also communicate these controls asynchronously with some delays. Algorithms of this type are discussed in generality in the book [BeT89], and also in the papers [BeY10], [BeY12], [YuB13], within the present DP context [see also the books [Ber12] (Section 2.6), and [Ber18] (Section 2.6)]. An example of such an algorithm is obtained when at a given stage, agent $\ell$ computes the rollout control $\tl u_k^\ell$ before knowing the rollout controls of some of the agents $1,\ldots,\ell-1$, and uses the controls $\m_k^1(x_k),\ldots,\m_k^{\ell-1}(x_k)$ of the base policy in their place. 
While such an algorithm is likely to work well for many problems, it may not possess the cost improvement property. In fact we can construct a simple example involving a single state, two agents, and two controls per agent, where the 2nd agent does not take into account the control applied by the 1st agent, and as a result the rollout policy performs worse than the base policy.

\xdef\examplecostdet{\exampl}\examplnum\show{myexample}

\beginexample{\examplecostdet\ (Cost Deterioration in the Absence of Adequate Agent Coordination)}Consider a problem with two agents ($m=2$) and a single state. Thus the state does not change and the costs of different stages are decoupled (the problem is essentially static). Each of the two agents has two controls: $u_k^1\in\{0,1\}$ and 
 $u_k^2\in\{0,1\}$. The cost per stage $g_k$ is equal to 0 if $u_k^1\ne u_k^2$, is equal to 1 if $u_k^1=u_k^2=0$, and  is equal to 2 if $u_k^1=u_k^2=1$. Suppose that the base policy applies $u_k^1=u_k^2=0$. Then it can be seen that when executing rollout, the first agent applies $u_k^1=1$, and in the absence of knowledge of this choice, the second agent  also applies $u_k^2=1$ (thinking that the first agent will use the base policy control $u_k^1=0$). Thus the cost of the rollout policy is 2 per stage, while the cost of the base policy is 1 per stage. By contrast the rollout algorithm that takes into account the first agent's control when selecting the second agent's control applies $u_k^1=1$ and $u_k^2=0$, thus resulting in a rollout policy with the optimal cost of 0 per stage. 
 
 The difficulty here is inadequate coordination between the two agents. In particular, each agent uses rollout to compute the local control, each thinking that the other will use the base policy control. If instead the two agents were to coordinate their control choices, they would have applied an optimal policy.
 \endexample
\vskip-0.5pc 

The simplicity of the preceding example also raises serious questions as to whether the cost improvement property \costimprstoch\ can be easily maintained by a distributed rollout algorithm where the agents do not know the controls applied by the preceding agents in the given order of local control selection, and use instead the controls of the base policy. Still, however, such an algorithm is computationally attractive in view of its potential for efficient distributed implementation, and may be worth considering in a practical setting. A noteworthy property of this algorithm is that if the base policy is optimal, the same is true of the rollout policy. This suggests that if the base policy is nearly optimal, the same is true of the rollout policy. 

 One may also speculate that if agents are naturally ``weakly coupled" in the sense that their choice of control has little impact in the desirability of various controls of other agents, then a more flexible inter-agent communication pattern may be sufficient for cost improvement.\footnote{\dag}{\ninepoint In particular, one may divide the agents in ``coupled" groups, and require coordination of control selection only within each group, while the computation of different groups may proceed in parallel. For example, in applications where the agents' locations are distributed within some geographical area, it may make sense to form agent groups on the basis of geographic proximity, i.e., one may require that agents that are geographically near each other  (and hence are more coupled) coordinate their control selections, while agents that are geographically far apart (and hence are less coupled) forego any coordination.} A computational comparison of various multiagent rollout algorithms with flexible communication patterns may shed some light on this question. The key question is whether and under what circumstances agent coordination is essential, i.e., there is a significant performance loss when the computations of different agents are done to some extent concurrently rather than sequentially with intermediate information exchange.

\vskip-1.5pc

\section{Multiagent Problem Formulation - Infinite Horizon Discounted\hfill\break Problems}

\xdef\figinfinitehorizon{\figr}\figrnum\show{myfigure}

\pn The multiagent rollout ideas that we have discussed so far can be modified and generalized to apply to infinite horizon problems. In this context, we may consider multiagent versions of value  iteration (VI for short) and policy iteration (PI for short) algorithms. We will focus on discounted problems with finite number of states and controls, so that the Bellman operator is a contraction mapping, and the strongest version of the available theory  applies (the solution of Bellman's equation is unique, and strong convergence results hold for VI and PI methods); see [Ber12], Chapters 1 and 2, and [Ber18], Chapter 2. However, a qualitatively similar methodology can be applied to undiscounted problems involving a termination state (e.g., stochastic shortest path problems, see [BeT96], Chapter 2, [Ber12], Chapter 3, and [Ber18], Chapters 3 and 4).

In particular, we consider a standard Markovian decision problem (MDP for short)  infinite horizon discounted version of the finite horizon $m$-agent problem of Section 1.2, where $m>1$.
The control $u$ consists of $m$ components $u_\ell$, $\ell=1,\ldots,m$,
$$u=(u_1,\ldots,u_m),$$
(for the MDP notation adopted for this section, we switch for convenience to subscript indexing for control components, and reserve superscript indexing for policy iterates). At state $x$ and stage $k$, a control $u$ is applied, and the system transitions to a next state $y$ with transition probabilities $p_{xy}(y)$ and cost $g(x,u,y)$. When at stage $k$ the transition cost is discounted by $\a^k$, where $\a\in(0,1)$ is the discount factor. 
Each control component $u_\ell$ is separately constrained to lie in a given finite set $U_\ell (x)$ when the system is at state $x$. 
Thus the control constraint is $u\in U(x)$, where $U(x)$ is the finite Cartesian product set
$$U(x)=U_1(x)\times\cdots\times U_m(x).$$
The cost function of a stationary policy $\m$ that applies control $\m(x)\in U(x)$ at state $x$ is denoted by $J_\m(x)$, and the optimal cost [the minimum over $\m$ of $J_\m(x)$] is denoted $\jstar(x)$.

\topinsert
\centerline{\hskip0pc\includegraphics[width=5.5in]{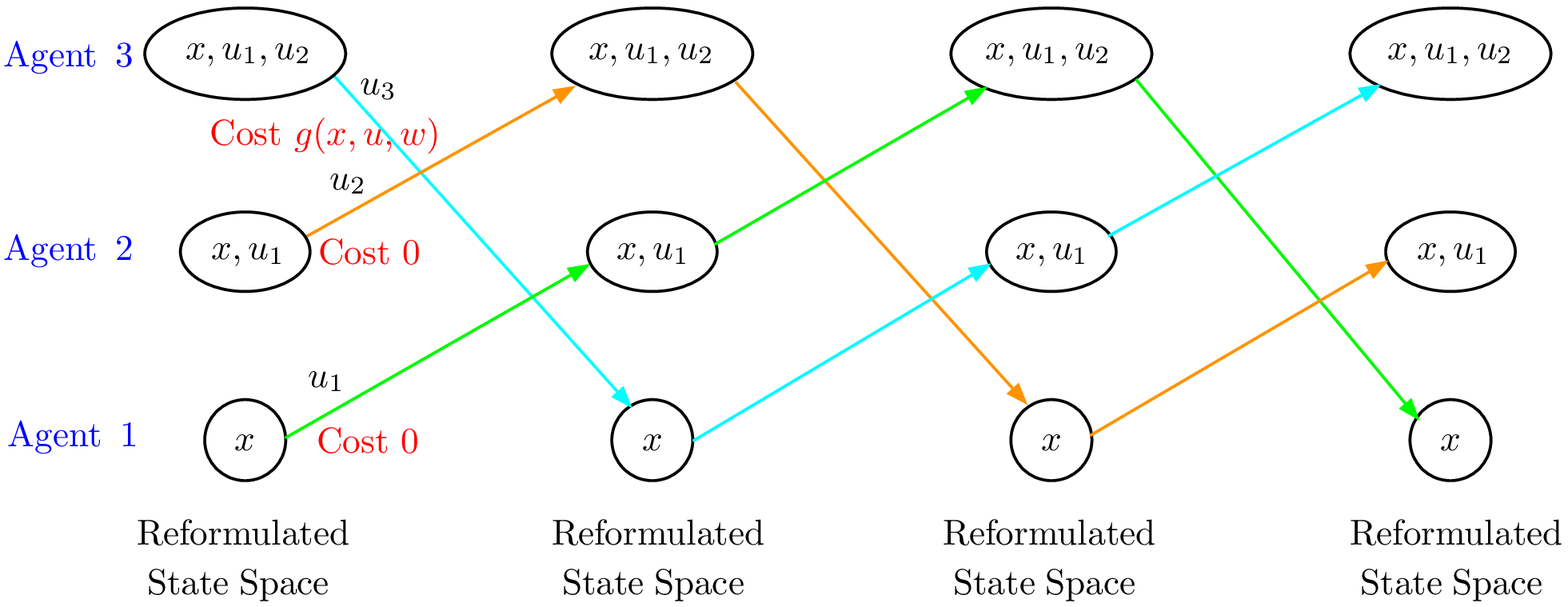}}\vskip0pc
\hskip-3pc\fig{-1.5pc}{\figinfinitehorizon} {Illustration of how to transform an $m$-agent infinite horizon problem into a stationary infinite horizon problem with fewer control choices available at each state (in this figure $m=3$). At the typical stage and state $x$, the first agent chooses $u_1$ at no cost leading to state $(x,u_1)$. Then the second agent applies $u_2$ at no cost leading to state $(x,u_1,u_2)$. Finally, the third agent applies $u_3$ leading to some state $y$ at cost $g(x,u,y)$, where $u$ is the combined control of the three agents, $u=(u_1,u_2,u_3)$. The figure shows the first three transitions of the trajectories that start from the states $x$, $(x,u_1)$, and $(x,u_1,u_2)$, respectively.}
\endinsert

An equivalent version of the problem, involving a reformulated/expanded state space is depicted in Fig.\ \figinfinitehorizon\ for the case $m=3$. The state space of the reformulated problem consists of
$$x,(x,u_1),\ldots,(x,u_1,\ldots,u_{m-1}),$$
where $x$ ranges over the original state space, and each $u_\ell$, $\ell=1,\ldots,m,$ ranges over the corresponding constraint set $U_\ell(x)$.
At each stage, the agents choose their controls sequentially in a fixed order: from state $x$ agent 1 applies $u_1\in U_1(x)$ to go to state $(x,u_1)$, then agent 2 applies $u_2\in U_2(x)$ to go to state  $(x,u_1,u_2)$, and so on, until finally at state $(x,u_1,\ldots,u_{m-1})$, agent $m$ applies $u_m\in U_m(x)$, completing the choice of control  $u=(u_1,\ldots,u_m)$, and effecting the transition to state $y$ at a cost $g(x,u,y)$, appropriately discounted. 

Note that this reformulation involves the type of tradeoff between control space complexity and state space complexity that we discussed in Section 1.3.
The reformulated problem involves $m$ cost-to-go functions 
$$J^0(x),J^1(x,u_1),\ldots,J^{m-1}(x,u_1,\ldots,u_{m-1}),\xdef\costfunctions{\lab}\eqnum\show{oneo}$$
with corresponding sets of Bellman equations, but a much smaller control space. Moreover, the existing analysis of rollout algorithms, including implementations, variations, and error bounds, applies to the reformulated problem; see Section 5.1 of the author's RL textbook [Ber19]. Similar to the finite horizon case, the implementation of the rollout algorithm involves one-agent-at-a-time policy improvement, and is much more economical for the reformulated problem, while maintaining the basic cost improvement and error bound properties of rollout, as they apply to the reformulated problem.

\subsection{Agent-by-Agent Policy Iteration for Infinite Horizon Discounted Problems}

\pn A PI algorithm generates a sequence of policies $\{\p^k\}$, and can be viewed as repeated or perpetual rollout, i.e.,  $\p^{k+1}$ is the rollout policy obtained when $\p^k$ is the base policy. 
For the reformulated problem just described, the  policy evaluation step of the PI algorithm requires the calculation of  $m$ cost-to-go functions of the form \costfunctions, so the policy evaluation must be done over a much larger space than the original state space. Moreover, the policies generated by this algorithm are also defined over a larger space and have the form
$$\m^0(x),\m^1(x,u^1),\ldots,\m^{m-1}(x,u^1,\ldots,u^{m-1}).\eqnum\show{oneo}$$
Motivated by this fact, we may consider a multiagent PI algorithm that operates over the simpler class of policies of the form
$$\m(x)=\big(\m^0(x),\m^1(x),\ldots,\m^{m-1}(x)\big),\eqnum\show{oneo}$$
i.e., the policies for the original infinite horizon problem.

We assume $n$ states, denoted by $1,\ldots,n$, and we  introduce the Bellman operator $T$, which maps a vector $J=\big(J(1),\ldots,J(n)\big)$ to the vector $TJ=\big((TJ)(1),\ldots,(TJ)(n)\big)$ according to
$$(TJ)(x)=\min_{u\in U(x)}\sum_{y=1}^np_{xy}(u)\big(g(x,u,y)+\a J(y)\big),\qquad x=1,\ldots,n,\xdef\tmap{\lab}\eqnum\show{oneo}$$
and for each policy $\m$, the corresponding Bellman operator $T_\m$ defined by
$$(T_\m J)(x)=\sum_{y=1}^np_{xy}\big(\m(x)\big)\Big(g\big(x,\m(x),y\big)+\a J(y)\Big),\qquad x=1,\ldots,n.\xdef\tmumap{\lab}\eqnum\show{oneo}$$
It is well known that $T$ and $T_\m$ are contraction mappings of modulus $\a$ with respect to the sup norm, and their unique fixed points are $\jstar$ and $J_\m$, respectively, i.e., $\jstar=T\jstar$ and $J_\m=T_\m J_\m$. 

In the preceding expressions, we will often expand $u$ to write it in terms of its components. In particular, we may write $p_{xy}(u_1,\ldots,u_m)$ and $g(x,u_1,\ldots,u_m,y)$ in place of $p_{xy}(u)$ and $g(x,u,y)$, respectively. Similarly, we will denote the components of a policy $\m$ as $\m_1,\ldots,\m_m$.

\subsubsection{The Standard Policy Iteration Algorithm}

\pn
For each policy $\m$, we introduce the subset of policies
$${\cal M}(\m)=\{\tl \m\mid T_{\tl \m} J_\m=TJ_\m\}.\xdef\calmu{\lab}\eqnum\show{oneo}$$
Equivalently, we have $\tl\m\in {\cal M}(\m)$ if $\tl \m$ is obtained from $\m$ by using the standard policy improvement operation,
$$\tl\m(x)\in\arg\min_{u\in U(x)}\sum_{y=1}^np_{xy}(u)\big(g(x,u,y)+\a J_\m(y)\big),\qquad x=1,\ldots,n,\xdef\polimprtove{\lab}\eqnum\show{oneo}$$
or 
$$\tl\m(x)\in\arg\min_{u\in U(x)}Q_\m(x,u),\qquad x=1,\ldots,n,\xdef\qfactorop{\lab}\eqnum\show{oneo}$$
where $Q_\m(x,u)$ is the Q-factor of the state-control pair $(x,u)$ corresponding to $\m$, given by
$$Q_\m(x,u)=\sum_{y=1}^np_{xy}(u)\big(g(x,u,y)+\a J_\m(y)\big).\xdef\qfactor{\lab}\eqnum\show{oneo}$$

The standard form of PI generates a sequence $\{\m^k\}$ of policies, starting from a given policy $\m^0$ (see e.g., [Ber12]). Given the current policy $\m^k$, it generates a new policy from the set of ``improved" policies ${\cal M}({\m^k})$ of Eq.\ \calmu:
$$\m^{k+1}\in {\cal M}({\m^k}).\eqnum\show{oneo}$$
Thus the $k$th iteration of the standard PI algorithm can be separated into two phases:
\nitem{(a)} {\it Policy evaluation\/}, which computes $J_{\m^k}$.
\nitem{(b)} {\it Policy improvement\/}, which computes a new policy $\m^{k+1}\in {\cal M}({\m^k})$ by the minimization over $u\in U(x)$ of the Q-factor $Q_{\m^k}(x,u)$; cf.\  Eq.\ \polimprtove-\qfactorop.
\smskip

\subsubsection{The Multiagent Policy Iteration Algorithm}

\pn Our proposed one-agent-at-a-time PI algorithm uses a modified form of policy improvement, whereby the control $u=(u_1,\ldots,u_m)$ is optimized one-component-at-a-time, similar to Section 2. In particular, given the current policy $\m^k$, the next policy is obtained as
$$\m^{k+1}\in \widetilde  {\cal M}({\m^k}),\xdef\incrpoliter{\lab}\eqnum\show{oneo}$$
where  for a given $\m$, we denote by $\widetilde {\cal M}({\m})$ the set of  policies $\tl \m=(\tl \m_1,\ldots,\tl \m_m)$ satisfying for all $x=1,\ldots,n,$
$$\tl \m_1(x)\in\arg\min_{u_1\in U_1(x)}\sum_{y=1}^np_{xy}\big(u_1,\m_2(x),\ldots,\m_m(x)\big)\Big(g\big(x,u_1,\m_2(x),\ldots,\m_m(x),y\big)+\a J_\m(y)\Big),$$
$$\tl \m_2(x)\in\arg\min_{u_2\in U_2(x)}\sum_{y=1}^np_{xy}\big(\tl\m_1(x),u_2,\m_3(x),\ldots,\m_m(x)\big)\Big(g\big(x,\tl\m_1(x),u_2,\m_3(x),\ldots,\m_m(x),y\big)+\a J_\m(y)\Big),$$
$$\cdots\qquad\qquad\cdots\qquad\qquad\cdots$$
$$\tl \m_m(x)\in\arg\min_{u_m\in U_m(x)}\sum_{y=1}^np_{xy}\big(\tl\m_1(x),\ldots,\tl\m_{m-1}(x),u_m\big)\Big(g\big(x,\tl\m_1(x),\ldots,\tl\m_{m-1}(x),u_m,y\big)+\a J_\m(y)\Big).\xdef\hatcalmu{\lab}\eqnum\show{oneo}$$

Note that each of the $m$ minimizations \hatcalmu\ can be performed for each state $x$ independently, i.e., the computations for state $x$ do not depend on the computations for other states, thus allowing the use of parallel computation over the different states. On the other hand, the computations corresponding to individual components must be performed in sequence (in the absence of special structure related to coupling of the control components through the transition probabilities and the cost per stage). It will also be clear from the subsequent analysis that the ordering of the components may change from one policy improvement operation to the next.

Similar to the finite horizon case of Sections 2 and 3, the salient feature of the one-agent-at-a-time policy improvement operation \hatcalmu\ is that it is far more economical than the standard policy improvement: it requires a sequence of $m$ minimizations, once over each of the control components $u_1,...,u_m$. In particular, for the minimization over the typical component $u_\ell$, the preceding components $u_1,\ldots,u_{\ell-1}$ have been computed earlier by the minimization that yielded the policy components $\tl\m_1,\ldots,\tl\m_{\ell-1}$, while the following controls $u_{\ell+1},\ldots,u_m$ are determined by the current policy components $\m_{\ell+1},\ldots,\m_m$. Thus, if the number of controls within each component constraint set $U_\ell(x)$ is bounded by a number $s$, the modified policy improvement phase requires at most $sm  n$ calculations of a Q-factor of the generic form \qfactor. 
By contrast, since the number of elements in the constraint set $U(x)$ is bounded by $s^m$, the corresponding number of Q-factor calculations in the standard policy improvement is bounded by $s^m  n$. Thus the one-agent-at-a-time policy improvement where the number of Q-factors grows linearly with $m$, as opposed to the standard policy improvement, where the number of Q-factor calculations grows exponentially with $m$.

We say that a policy $\m=\{\m_1,\ldots,\m_m\}$ is {\it agent-by-agent optimal} if it satisfies $\m\in \widetilde  {\cal M}(\m)$, or equivalently,  for all $x=1,\ldots,n$, and $\ell=1,\ldots,m$, we have 
$$\eqalign{\sum_{y=1}^np_{xy}\big(\m_1(x),\ldots,\m_m(x)\big)&\Big(g\big(x,\m_1(x),\ldots,\m_m(x),y\big)+\a J_{\m}(y)\Big)\cr
&=\min_{u_\ell\in U_\ell(x)}\sum_{y=1}^n  p_{xy}\big(\m_1(x),\ldots,\m_{\ell-1}(x),u_\ell,\m_{\ell+1}(y),\ldots,\m_{m}(y)\big)\cr
&\ \ \ \ \ \ \ \ \ \ \ \ \ \ \ \ \ \ \ \ \ \  \cdot \Big(g\big(\m_1(y),\ldots,\m_{\ell-1}(y),u_\ell,\m_{\ell+1}(y),\ldots,\m_{m}(y)\big)+\a J_{\m}(y)\Big).\cr}\xdef\agentbyagent{\lab}\eqnum\show{oneo}$$ 
To interpret this definition, let a policy $\m=\{\m_1,\ldots,\m_m\}$ be given, and consider for every  $\ell=1,\ldots,m$ the single agent DP problem where for all $\ell'\ne \ell$ the $\ell'$th policy  components  are fixed at $\m_{\ell'}$, while the $\ell$th policy component is subject to optimization. We view the   agent-by-agent optimality definition as the optimality condition for all the single agent problems [Eq.\ \agentbyagent\ can be written as $T_{\m,\ell} J_\m=T_\ell J_\m$, where $T_\ell$ and $T_{\m,\ell}$ are the Bellman operators \tmap\ and \tmumap\ that correspond to the single agent problem of agent $\ell$].  We can then conclude that $\m=\{\m_1,\ldots,\m_m\}$ is agent-by-agent optimal if each component $\m_\ell$ is optimal for the $\ell$th single agent problem, where it is assumed that the remaining policy components remain fixed; in other words by using $\m_\ell$, each agent $\ell$ acts optimally, assuming all other agents $\ell'\ne \ell$ use the corresponding policy components $\m_{\ell'}$. In the terminology of team theory such a policy may also be called ``person-by-person optimal."

Note that an (overall) optimal policy is agent-by-agent optimal, but the reverse is not true as the following example shows.  This is well-known from the aforementioned research on team theory; see [Mar55], [Rad62],  [Wit71],  [Ho80], [NMT13], [NaT19], [LTZ19], [ZPB19]. 

\xdef\examplecounterex{\exampl}\examplnum\show{myexample}

\beginexample{\examplecounterex\ (Counterexample for Agent-by-Agent Optimality)}Consider an infinite horizon problem, which involves two agents ($m=2$) and a single state $x$. Thus the state does not change and the costs of different stages are decoupled (the problem is essentially static). Each of the two agents has two controls: $u^1\in\{0,1\}$ and 
 $u^2\in\{0,1\}$. The cost per stage $g$ is equal to 2 if $u^1\ne u^2$, is equal to 1 if $u^1=u^2=0$, and  is equal to 0 if $u^1=u^2=1$. The unique optimal policy is to apply $\m_1(x)=1$ and $\m_2(x)=1$. However, it can be seen that the suboptimal policy that applies $\m_1(x)=0$ and $\m_2(x)=0$ is agent-by-agent optimal.
\endexample 
  
The preceding example is representative of an entire class of DP problems where an agent-by-agent optimal policy is not overall optimal. Any static multivariable optimization problem where there are nonoptimal solutions that cannot be improved upon by coordinate descent can be turned into an infinite horizon DP example where these nonoptimal solutions define agent-by-agent optimal policies that are not overall optimal. Conversely, one may search for problem classes where an agent-by-agent optimal policy is guaranteed to be (overall) optimal among the type of multivariable optimization problems where coordinate descent is guaranteed to converge to an optimal solution; for example positive definite quadratic problems or problems involving differentiable strictly convex functions (see [Ber16], Section 3.7). Generally, agent-by-agent optimality may be viewed as an acceptable form of optimality for many types of problems.

Our main result is that the agent-by-agent PI algorithm just described converges to an  agent-by-agent optimal policy in a finite number of iterations.
For the proof, we use a special rule for breaking ties in the policy improvement operation in favor of the current policy component. This rule is easy to enforce, and guarantees that the algorithm cannot cycle between policies. Without this tie-breaking rule, the following proof can be modified to show that while the generated policies may cycle, the corresponding cost function values converge to the cost function value of some agent-by-agent optimal policy.

\xdef\propmultiagentpoliter{\propn}\propnum\show{myproposition}

\texshopbox{\proposition{\propmultiagentpoliter: (PI Convergence to an Agent-by-Agent Optimal Policy)}Let $\{\m^k\}$ be a sequence generated by the agent-by-agent PI algorithm \incrpoliter\ assuming that ties in the policy improvement operation of  Eq.\ \hatcalmu\ are broken as follows: If for any $\ell=1,\ldots,m$ and $x=1,\ldots,n$, the control component $\mu_\ell(x)$ attains the 
minimum in Eq.\ \hatcalmu, we choose $\tl \mu_\ell(x)=\mu_\ell(x)$ [even if there are other control components within $U_\ell(x)$ that 
attain the minimum in addition to $\mu_\ell(x)$]. Then for all $x$ and $k$, we have 
$$J_{\m^{k+1}}(x)\le J_{\m^k}(x),$$
and after a finite number of iterations, we have ${\m^{k+1}}= {\m^k}$, in which case the policies $\m^{k+1}$ and $\m^k$ are agent-by-agent optimal.
}

\proof In the following proof and later all vector inequalities are meant to be componentwise, i.e., for any two vectors $J$ and $J'$, we write $J\le J'$ if $J(x)=J'(x)$ for all $x$.
The critical step of the proof is the following monotone decrease inequality:
$$T_{\tl \m}J\le T_{\m}J\le J,\qquad \hbox{for all $\tl \m\in  \widetilde {\cal M}(\m)$ and $J$ with $T_\m J\le J$},\xdef\monotonedecrease{\lab}\eqnum\show{oneo}$$
which yields as a special case $T_{\tl \m}J_\m\le J_\m$, since $T_{\m}J_\m= J_\m$. This parallels a key inequality for standard PI, 
namely that $T_{\tl \m}J_\m\le J_\m$, for all $\tl \m\in  {\cal M}(\m),$
which lies at the  heart of its convergence proof.
Once Eq.\ \monotonedecrease\ is shown, the  monotonicity of the operator $T_{\tl \m}$ implies the cost improvement property $J_{\tl \m}\le J_\m$, and by using the finiteness of the set of policies, the finite convergence of the algorithm will follow. 

We will give the proof of the monotone decrease inequality \monotonedecrease\ for the case  $m=2$. The proof for an arbitrary number of components $m>2$ is entirely similar. 
Indeed, if $T_\m J\le J$, we have for all $x$,
$$\eqalign{(T_{\tl \m}J)(x)&=\sum_{y=1}^np_{xy}\big(\tl \m_1(x),\tl \m_2(x)\big)\Big(g\big(x,\tl \m_1(x),\tl \m_2(x),y\big)+\a J(y)\Big)\cr
&=\min_{u_2\in U_2(x)}\sum_{y=1}^np_{xy}\big(\tl \m_1(x),u_2\big)\Big(g\big(x,\tl \m_1(x),u_2,y\big)+\a J(y)\Big)\cr
&\le \sum_{y=1}^np_{xy}\big(\tl \m_1(x),\m_2(x)\big)\Big(g\big(x,\tl \m_1(x),\m_2(x),y\big)+\a J(y)\Big)\cr
&=\min_{u_1\in U_1(x)}\sum_{y=1}^n  p_{xy}\big(u_1,\m_2(x)\big) \Big(g\big(x,u_1,\m_2(x),y\big)+\a J(y)\Big)\cr
&\le \sum_{y=1}^np_{xy}\big(\m_1(x),\m_2(x)\big)\Big(g\big(x,\m_1(x),\m_2(x),y\big)+\a J(y)\Big)\cr
&=(T_\m J)(x)\cr
&\le J(x),\cr}\xdef\decineqproof{\lab}\eqnum\show{oneo}$$
where:
\nitem{(1)} The first and fourth equalities use the definition of the Bellman operator $T_{\tl \m}$.
\nitem{(2)} The second and third equalities hold by the definition of policies $\tl \m\in \widetilde {\cal M}(\m)$.
\nitem{(3)} The first  and second inequalities are evident.
\nitem{(4)} The last inequality is the assumption $T_\m J\le J$.
\smskip


By letting $J=J_{\m^k}$ in the monotone decrease inequality \monotonedecrease, we have $T_{\m^{k+1}}J_{\m^k}\le J_{\m^k}$. In view of the monotonicity of $T_{\m^{k+1}}$, we also have $T_{\m^{k+1}}^{t+1}J_{\m^k}\le T_{\m^{k+1}}^t J_{\m^k}$ for all $t\ge1$, so that 
$$J_{\m^{k+1}}=\lim_{t\to\infty}T_{\m^{k+1}}^tJ_{\m^k}\le T_{\m^{k+1}}J_{\m^k}\le J_{\m^k}.$$
It follows that either $J_{\m^{k+1}}= J_{\m^k}$, or else we have 
strict policy improvement, i.e., $J_{\m^{k+1}}(x)< J_{\m^k}(x)$ for at least one state $x$. As long as strict improvement occurs, no generated policy can be repeated by the algorithm. Since there are only finitely many policies, it follows that within a finite number of iterations, we will have $J_{\m^{k+1}}= J_{\m^k}$. Once this happens, equality will hold throughout in Eq.\ \decineqproof\ when $\m=\m^k$, $\tl \m=\m^{k+1}$, and $J=J_{\m^k}$. This implies that
$$\eqalign{\sum_{y=1}^np_{xy}\big(\m^{k+1}_1(x),\m^{k+1}_2(x)\big)&\Big(g\big(x,\m^{k+1}_1(x),\m^{k+1}_2(x),y\big)+\a J_{\m^k}(y)\Big)\cr
&=\min_{u_2\in U_2(x)}\sum_{y=1}^np_{xy}\big(\m^{k+1}_1(x),u_2\big)\Big(g\big(x,\m^{k+1}_1(x),u_2,y\big)+\a J_{\m^k}(y)\Big)\cr
&=\sum_{y=1}^np_{xy}\big(\m^{k+1}_1(x),\m^k_2(x)\big)\Big(g\big(x,\m^{k+1}_1(x),\m^k_2(x),y\big)+\a J_{\m^k}(y)\Big),\cr}\xdef\decineqproofo{\lab}\eqnum\show{oneo}$$
and 
$$\eqalign{\sum_{y=1}^np_{xy}\big(\m^{k+1}_1(x),\m^k_2(x)\big)&\Big(g\big(x,\m^{k+1}_1(x),\m^k_2(x),y\big)+\a J_{\m^k}(y)\Big)\cr
&=\min_{u_1\in U_1(x)}\sum_{y=1}^n  p_{xy}\big(u_1,\m^k_2(x)\big) \Big(g\big(x,u_1,\m^k_2(x),y\big)+\a J_{\m^k}(y)\Big)\cr
&= \sum_{y=1}^np_{xy}\big(\m^k_1(x),\m^k_2(x)\big)\Big(g\big(x,\m^k_1(x),\m^k_2(x),y\big)+\a J_{\m^k}(y)\Big).\cr
}\xdef\decineqprooft{\lab}\eqnum\show{oneo}
$$
In view of our tie breaking rule, Eq.\ \decineqprooft\ implies that $\m_1^{k+1}=\m_1^k$, and then Eq.\ \decineqproofo\ implies that $\m_2^{k+1}=\m_2^k$. Thus we have $\m^{k+1}=\m^k$, and from Eqs.\ \decineqproofo\ and \decineqprooft, $\m^{k+1}$ and $\m^k$ are agent-by-agent optimal. \qed

As Example \examplecounterex\ shows, there may be multiple agent-by-agent optimal policies, with different cost functions. This illustrates that the policy obtained by the multiagent PI  algorithm may depend on the starting policy. It turns out that the same example can be used to show that the policy obtained by the algorithm depends also on the order in which the agents select their controls.

\xdef\examplecounterext{\exampl}\examplnum\show{myexample}

\beginexample{\examplecounterext\ (Dependence of the Final Policy on the Agent Iteration Order)}Consider the problem of Example \examplecounterex. In this problem there are two agent-by-agent optimal policies: the optimal policy $\m^*$ where $\m^*_1(x)=1$ and $\m^*_2(x)=1$,  and the suboptimal policy $\hat \m$ where $\hat \m_1(x)=0$ and $\hat \m_2(x)=0$. Let the starting policy be $\m^0$ where $\m_1^0(x)=1$ and $\m_2^0(x)=0$. Then if agent 1 iterates first, the algorithm will terminate with the suboptimal policy, $\m^1= \hat \m$, while if agent 2 iterates first, the algorithm will terminate with the optimal policy, $\m^1= \m^*$.
\endexample 

Generally, it may not be easy to escape from a suboptimal agent-by-agent optimal policy. This is similar to trying to escape from a local minimum in multivariable optimization. Of course, one may try minimization over suitable subsets of control components, selected by some heuristic, possibly randomized, mechanism. However, such a approach is likely to be problem-dependent, and may not offer meaningful guarantees of success.

We note that the line of proof based on the monotone decrease inequality \monotonedecrease\ given above can be used to establish the validity of some variants of agent-by-agent PI. One such variant, which we will not pursue further, enlarges the set  $\widetilde  {\cal M}(\m)$ to allow approximate minimization over the control components in Eq.\ \hatcalmu. In particular, we require that in place of Eq.\ \decineqproof, each control $\tl \m_\ell (x)$, $\ell=1,\ldots,m$, satisfies 
$$\eqalign{\sum_{y=1}^np_{xy}&\big(\tl\m_1(x),\ldots,\tl\m_{\ell-1}(x),\tl\m_\ell(x),\m_{\ell+1}(x),\ldots,\m_m(x)\big)\cr
&\ \ \ \ \ \ \ \ \ \ \ \ \ \ \ \Big(g\big(x,\tl\m_1(x),\ldots,\tl\m_{\ell-1}(x),\tl\m_\ell(x),\m_{\ell+1}(x),\ldots,\m_m(x),y\big)+
\a J_\m(y)\Big)\cr
&<\sum_{y=1}^np_{xy}\big(\tl\m_1(x),\ldots,\tl\m_{\ell-1}(x),\m_\ell(x),\m_{\ell+1}(x),\ldots,\m_m(x)\big)\cr
&\ \ \ \ \ \ \ \ \ \ \ \ \ \ \ \Big(g\big(x,\tl\m_1(x),\ldots,\tl\m_{\ell-1}(x),\m_\ell(x),\m_{\ell+1}(x),\ldots,\m_m(x),y\big)+\a J_\m(y)\Big),\cr}$$
whenever there exists $u_\ell\in U_\ell(x)$  that can strictly reduce the corresponding minimized expression in Eq.\ \decineqproof. It can be seen that even with this approximate type of minimization over control components, the convergence proof of Prop.\ \propmultiagentpoliter\ still goes through.

Another  important variant of agent-by-agent PI  is an optimistic version, whereby policy evaluation is performed by using a finite number of agent-by-agent value iterations. Moreover, there are many possibilities for approximate agent-by-agent PI versions, including the use of value and policy neural networks. 
In particular, the multiagent policy improvement operation \hatcalmu\ may be performed at a sample set of states $x^s$, $s=1,\ldots,q$, thus yielding a training set of state-rollout control pairs $\big(x^s,\tl \m(x^s)\big)$, $s=1,\ldots,q$, which can be used to train a (policy) neural network to generate an approximation $\hat \m$ to the policy $\tl \m$. The policy $\hat \m$ can be used in turn to train a feature-based architecture or a neural network that approximates its cost function $J_{\hat \m}$, and the approximate multiagent PI cycle can be continued. Thus in this scheme, the difficulty with a large control space is mitigated by agent-by-agent policy improvement, while the difficulty with a large state space is overcome by training value and policy networks. A further discussion of this type of approximate schemes is beyond the scope of the present paper.

Finally, we note that the issues relating to parallelization of the policy improvement (or rollout) step that we discussed at the end of Section 3 for finite horizon problems, also apply to infinite horizon problems. Moreover, the natural partition of the state space illustrated in Fig.\ \figinfinitehorizon\ suggests a distributed implementation (which may be independent of any parallelization in the policy improvement step). In particular, distributed asynchronous PI algorithms based on state space partitions are proposed and analyzed in the work of Bertsekas and Yu [BeY10]  [see also [BeY12], [YuB13], and the books [Ber12] (Section 2.6), and [Ber18] (Section 2.6)]. These algorithms are  relevant for distributed implementation of the multiagent PI ideas of the present paper. 

\vskip-1pc

\section{Concluding Remarks}
\vskip-0.9pc
\pn We have shown that in the  context of multiagent problems, an agent-by-agent version of the rollout algorithm has greatly reduced computational requirements, while still maintaining the fundamental cost improvement property of the standard rollout algorithm. There are many variations of rollout algorithms for multiagent problems, which deserve attention, despite the potential lack of strict cost improvement in the case of a suboptimal base policy that is agent-by-agent optimal. Computational tests in some practical multiagent settings will be helpful in comparatively evaluating some of these variations.

We have primarily focused on the cost improvement property, and the practically important fact that it can be achieved at a much reduced computational cost. However, it is useful to keep in mind that the agent-by-agent rollout algorithm is simply the standard all-agents-at-once rollout algorithm applied to the (equivalent) reformulated problem of Fig.\ \figunfolded\ (or Fig.\ \figinfinitehorizon\ in the infinite horizon case). As a result, all known insights, results, error bounds, and approximation techniques for standard rollout apply in suitably reformulated form. 

In this paper, we have assumed that the control constraint set is finite in order to argue about the computational efficiency of the agent-by-agent rollout algorithm. The rollout algorithm itself and its cost improvement property are valid even in the case where the control constraint set is infinite, including the model predictive control context (cf.\ Section 2.5 of the RL book [Ber19]), and linear-quadratic problems. However, it may be unclear that agent-by-agent rollout offers an advantage in the infinite control space case.

We have also discussed an agent-by-agent version of PI for infinite horizon problems, which uses one-component-at-a-time policy improvement. While this algorithm may terminate with a suboptimal policy that is agent-by-agent optimal, it may produce comparable performance to the standard PI algorithm, which however may be computationally intractable even for a moderate number of agents. Moreover, our multiagent PI convergence result of Prop.\ \propmultiagentpoliter\ can be extended beyond the finite-state discounted context to more general infinite horizon DP contexts, where the PI algorithm is well-suited for algorithmic solution. Other extensions include  agent-by-agent variants of VI, optimistic PI, and other related methods. The analysis of such extensions is reported separately; see [Ber20a].

We finally mention that the idea of agent-by-agent rollout also applies within the context of challenging deterministic discrete/combinatorial optimization problems, which involve constraints that couple the controls of different stages. We discuss the corresponding constrained multiagent rollout algorithms separately in the paper [Ber20b]. 

\vskip-1.5pc

\section{References}
\vskip-0.9pc
\def\ref{\vskip1.pt\pn}

\ref [BeT89]  Bertsekas, D.\ P., and Tsitsiklis, J.\ N., 1989.\ Parallel and Distributed Computation: Numerical Methods, Prentice-Hall, Englewood Cliffs, NJ; republished in 1996 by Athena Scientific, Belmont, MA.

\ref [BeT96]  Bertsekas, D.\ P., and Tsitsiklis, J.\ N., 1996.\ Neuro-Dynamic
Programming, Athena Scientific, Belmont, MA.

\ref[BeY10] Bertsekas, D.\ P., and Yu, H., 2010.\ ``Asynchronous Distributed Policy Iteration in Dynamic Programming,"  Proc.\ of Allerton Conf.\ on Communication, Control and Computing,  Allerton Park, Ill, pp.\ 1368-1374.

\ref[BeY12] Bertsekas, D.\ P., and Yu, H., 2012.\ ``Q-Learning and Enhanced Policy Iteration in Discounted 
Dynamic Programming,"  Math.\ of OR, Vol.\ 37, pp.\ 66-94.

\ref[Ber12] Bertsekas, D.\ P., 2012.\ Dynamic Programming and Optimal Control, Vol.\ II, 4th edition, Athena Scientific, Belmont, MA.

\ref[Ber16] Bertsekas, D.\ P., 2016.\ Nonlinear Programming, 3rd edition, Athena Scientific, Belmont, MA.

\ref[Ber17] Bertsekas, D.\ P., 2017.\ Dynamic Programming and Optimal Control, Vol.\ I, 4th edition, Athena Scientific, Belmont, MA.

\ref[Ber18] Bertsekas, D.\ P., 2018.\ Abstract Dynamic Programming, Athena Scientific, Belmont, MA.

\ref [Ber19]  Bertsekas, D.\ P., 2019.\ Reinforcement Learning and Optimal Control, Athena Scientific, Belmont, MA.

\ref[Ber20a] Bertsekas, D.\ P., 2020.\ ``Multiagent Value Iteration Algorithms in Dynamic Programming and Reinforcement Learning," in preparation.

\ref[Ber20b] Bertsekas, D.\ P., 2020.\ ``Constrained Multiagent Rollout and Multidimensional Assignment with the Auction Algorithm," arXiv preprint, arXiv:2002.07407.

\ref[Ho80] Ho, Y.\ C., 1980.\ ``Team Decision Theory and Information Structures," Proceedings of the IEEE, Vol.\ 68, pp.\ 644-654.

\ref[LTZ19] Li, Y., Tang, Y., Zhang, R., and Li, N., 2019.\ ``Distributed Reinforcement Learning for Decentralized Linear Quadratic Control: A Derivative-Free Policy Optimization Approach," arXiv preprint arXiv:1912.09135.

\ref[Mar55] Marschak, J., 1975.\ ``Elements for a Theory of Teams," Management Science, Vol.\ 1, pp.\ 127-137.

\ref[NMT13] Nayyar, A., Mahajan, A. and Teneketzis, D., 2013.\ ``Decentralized Stochastic Control with Partial History Sharing: A Common Information Approach," IEEE Transactions on Automatic Control, Vol.\ 58, pp.\ 1644-1658.

\ref[NaT19] Nayyar, A. and Teneketzis, D., 2019.\ ``Common Knowledge and Sequential Team Problems," IEEE Transactions on Automatic Control, Vol.\ 64, pp.\ 5108-5115.

\ref[Rad62] Radner, R., 1962.\ ``Team Decision Problems," Ann.\ Math.\ Statist., Vol.\ 33, pp.\ 857-881.

\old{
\ref[WaS00] de Waal, P.\ R., and van Schuppen, J.\ H., 2000.\ ``A Class of Team Problems with Discrete Action Spaces: Optimality Conditions Based on Multimodularity," SIAM J.\ on Control and Optimization, Vol.\ 38, pp.\ 875-892.
}

\ref[Wit71] Witsenhausen, H., 1971.\ ``Separation of Estimation and Control for Discrete Time Systems," Proceedings of the IEEE, Vol.\ 59, pp.\ 1557-1566.

\ref[YuB13] Yu, H., and Bertsekas, D.\ P., 2013.\ ``Q-Learning and Policy Iteration Algorithms for Stochastic Shortest Path Problems," Annals of Operations Research, Vol.\ 208, pp.\ 95-132.

\ref[ZPB19] Zoppoli, R., Parisini, T., Baglietto, M., and Sanguineti, M., 2019.\ Neural Approximations for Optimal Control and Decision, Springer.

\end